\documentclass[sigconf]{acmart}
\usepackage{xcolor,colortbl}
\usepackage{xspace}
\usepackage{rotating}
\usepackage{booktabs}
\usepackage{makecell}
\usepackage{amsmath}
\usepackage{wrapfig}
\usepackage{multirow}
\usepackage{graphicx}
\usepackage{graphbox}
\usepackage{tabularray}
\usepackage[export]{adjustbox}
\usepackage{xr}
\usepackage{enumitem}
\usepackage[abbreviations]{glossaries-extra}

\newcommand{\etal}{~et al.\@\xspace}

\newcommand{\eg}{e.g.\@\xspace}

\newcommand{\ie}{i.e.\@\xspace}

\newcommand{\refSec}[1]{Sec.~\ref{sec:#1}}
\newcommand{\refFig}[1]{Fig.~\ref{fig:#1}}

\newcommand{\refEq}[1]{Eq.~(\ref{eq:#1})}
\newcommand{\refTbl}[1]{Tbl.~\ref{tbl:#1}}

\newabbreviation{CNN}{CNN}{Convolutional Neural Network}
\newabbreviation{VR}{VR}{Virtual Reality}
\newabbreviation{AR}{AR}{Augmented Reality}
\newabbreviation{GAN}{GAN}{Generative Adversarial Network}
\newabbreviation{MTL}{MTL}{Multitask Learning}
\newabbreviation{MLP}{MLP}{Multilayer Perceptron}
\newabbreviation{DNN}{DNN}{Deep Neural Network}
\newabbreviation{HVS}{HVS}{Human Visual System}
\newabbreviation{PSNR}{PSNR}{Peak Signal-to-Noise-Ratio}
\newabbreviation{SSIM}{SSIM}{Structural Similarity Index Measure}
\newabbreviation{LPIPS}{LPIPS}{Learned Perceptual Image Patch Similarity Metric}

\global\long\def\VR{\gls{VR}\xspace}
\global\long\def\AR{\gls{AR}\xspace}
\global\long\def\GAN{\gls{GAN}\xspace}
\global\long\def\MTL{\gls{MTL}\xspace}
\global\long\def\MLP{\gls{MLP}\xspace}

\global\long\def\HVS{\gls{HVS}\xspace}
\global\long\def\PSNR{\gls{PSNR}\xspace}
\global\long\def\SSIM{\gls{SSIM}\xspace}
\global\long\def\LPIPS{\gls{LPIPS}\xspace}
\global\long\def\FovVideoVDP{FovVideoVDP\xspace}

\definecolor{Red}{rgb}{1.0, 0.0, 0.0}
\definecolor{Green}{rgb}{0., 0.6, 0.}
\definecolor{Yellow}{rgb}{0.9, 0.7, 0.0}

\newcommand{\high}[1]{{\cellcolor{green!25} #1}}
\newcommand{\medium}[1]{{\cellcolor{yellow!25} #1}}
\newcommand{\low}[1]{{\cellcolor{red!25} #1}}

\definecolor{table_shade1}{HTML}{D7D9EE}
\definecolor{table_shade2}{HTML}{CACCE9}
\definecolor{table_shade3}{HTML}{ffebe6}
\definecolor{table_shade4}{HTML}{ffd6cc}

\newcommand{\codebase}{\textcolor{blue}{\url{https://complightlab.com/multitasking_perceptual_graphics}}\@\xspace}

\AtBeginDocument{%
  }

\copyrightyear{2025}
\acmYear{2025}
\setcopyright{cc}
\setcctype{by}
\acmConference[MM '25]{Proceedings of the 33rd ACM International Conference on Multimedia}{October 27--31, 2025}{Dublin, Ireland}
\acmBooktitle{Proceedings of the 33rd ACM International Conference on Multimedia (MM '25), October 27--31, 2025, Dublin, Ireland}\acmDOI{10.1145/3746027.3754801}
\acmISBN{979-8-4007-2035-2/2025/10}

\begin{document}

%%
%% The "title" command has an optional parameter,
%% allowing the author to define a "short title" to be used in page headers.
\title{Learned Single-Pass Multitasking Perceptual Graphics for Immersive Displays}

%%
%% The "author" command and its associated commands are used to define
%% the authors and their affiliations.
%% Of note is the shared affiliation of the first two authors, and the
%% "authornote" and "authornotemark" commands
%% used to denote shared contribution to the research.
\author{Doğa Yılmaz}
\authornote{Doğa Yılmaz is the corresponding author.}
\orcid{0000-0002-2268-7136}
\affiliation{%
  \institution{University College London}
  \department{Department of Computer Science}
  \streetaddress{169 Euston Road}
  \city{London}
  \country{United Kingdom}
  \postcode{NW1 2AE}
}
\email{doga.yilmaz.24@ucl.ac.uk}

\author{He Wang}
\orcid{0000-0002-2281-5679}
\affiliation{%
  \institution{University College London}
  \department{AI Center \\ Department of Computer Science}
  \streetaddress{169 Euston Road}
  \city{London}
  \country{United Kingdom}
  \postcode{NW1 2AE}
}
\email{he_wang@ucl.ac.uk}

\author{Towaki Takikawa}
\orcid{0000-0003-2019-1564}
\affiliation{%
  \institution{University of Toronto}
  \department{Department of Computer Science}
  \streetaddress{40 St George St}
  \city{Toronto, Ontario}
  \country{Canada}
}
\email{tovacinni@gmail.com}

\author{Duygu Ceylan}
\orcid{0000-0002-2307-9052}
\affiliation{%
  \institution{Adobe Research}
  \streetaddress{1 Old Street Yard}
  \city{London}
  \country{United Kingdom}
  \postcode{EC1Y 8AF}
}
\email{ceylan@adobe.com}

\author{Kaan Akşit}
\orcid{0000-0002-5934-5500}
\affiliation{%
  \institution{University College London}
  \department{Department of Computer Science}
  \streetaddress{169 Euston Road}
  \city{London}
  \country{United Kingdom}
  \postcode{NW1 2AE}
}
\email{k.aksit@ucl.ac.uk}

%%
%% By default, the full list of authors will be used in the page
%% headers. Often, this list is too long, and will overlap
%% other information printed in the page headers. This command allows
%% the author to define a more concise list
%% of authors' names for this purpose.
%% \renewcommand{\shortauthors}{Yilmaz et al.}
\renewcommand{\shortauthors}{Doğa Yılmaz, He Wang, Towaki Takikawa, Duygu Ceylan, and Kaan Akşit}

%%
%% The abstract is a short summary of the work to be presented in the
%% article.
\begin{abstract}
Emerging immersive display technologies efficiently utilize resources with perceptual graphics methods such as foveated rendering and denoising.
Running multiple perceptual graphics methods challenges devices with limited power and computational resources.
We propose a computationally-lightweight learned multitasking perceptual graphics model.
Given RGB images and text-prompts, our model performs text-described perceptual tasks in a single inference step.
Simply daisy-chaining multiple models or training dedicated models can lead to model management issues and exhaust computational resources.
In contrast, our flexible method unlocks consistent high quality perceptual effects with reasonable compute, supporting various permutations at varied intensities using adjectives in text prompts (\eg ``mildly'', ``lightly'').
Text-guidance provides ease of use for dynamic requirements such as creative processes.
To train our model, we propose a dataset containing source and perceptually enhanced images with corresponding text prompts.
We evaluate our model on desktop and embedded platforms and validate perceptual quality through a user study.
\end{abstract}

%%
%% The code below is generated by the tool at http://dl.acm.org/ccs.cfm.
%% Please copy and paste the code instead of the example below.
%%
\begin{CCSXML}
<ccs2012>
   <concept>
       <concept_id>10010147.10010371.10010387.10010393</concept_id>
       <concept_desc>Computing methodologies~Perception</concept_desc>
       <concept_significance>500</concept_significance>
       </concept>
    <concept>
       <concept_id>10010147.10010371.10010382</concept_id>
       <concept_desc>Computing methodologies~Image manipulation</concept_desc>
       <concept_significance>500</concept_significance>
       </concept>
   <concept>
       <concept_id>10010147.10010257.10010293.10010294</concept_id>
       <concept_desc>Computing methodologies~Neural networks</concept_desc>
       <concept_significance>300</concept_significance>
    </concept>
 </ccs2012>
\end{CCSXML}

\ccsdesc[500]{Computing methodologies~Perception}
\ccsdesc[500]{Computing methodologies~Image manipulation}
\ccsdesc[300]{Computing methodologies~Neural networks}

%%
%% Keywords. The author(s) should pick words that accurately describe
%% the work being presented. Separate the keywords with commas.
\keywords{Perceptual Graphics, Immersive Displays, Generative Multimedia}
%% A "teaser" image appears between the author and affiliation
%% information and the body of the document, and typically spans the
%% page.
\begin{teaserfigure}
\includegraphics[width=\textwidth]{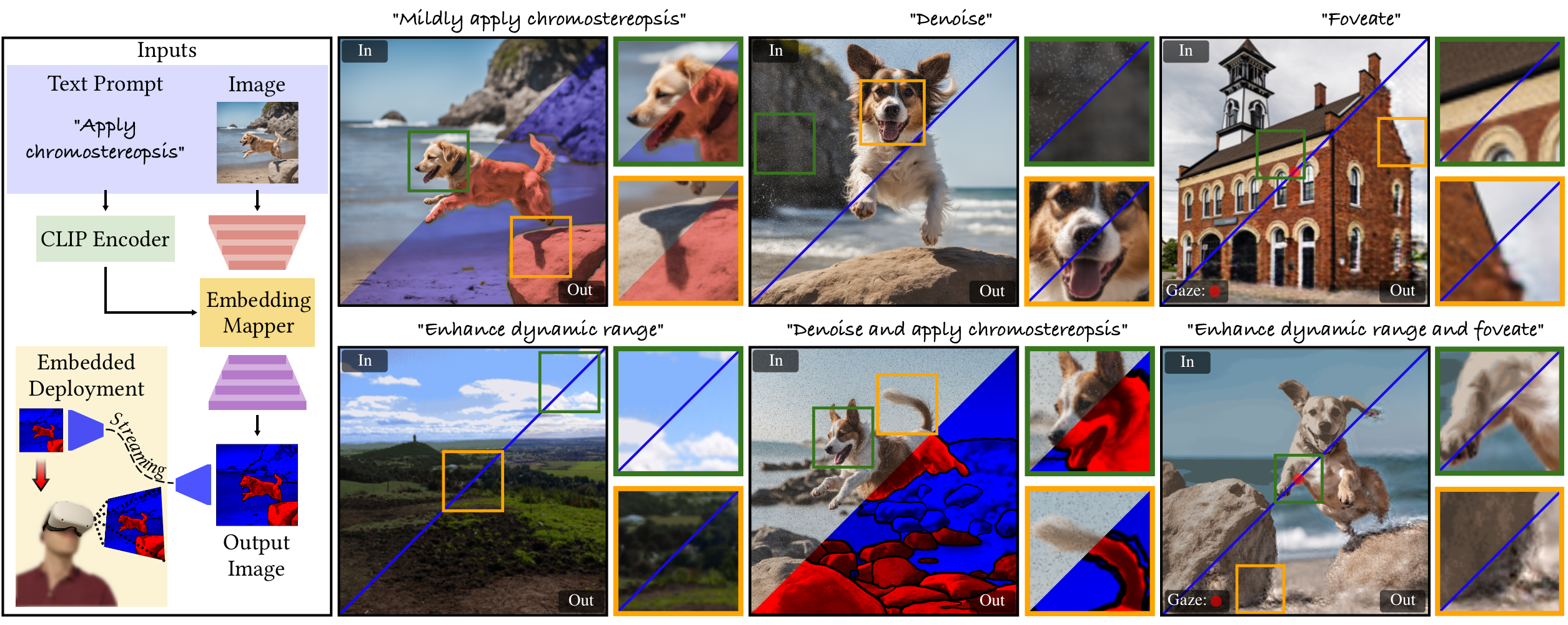}
\caption{
	 Our novel text-guided model (left) perceptually enhances input images based on given prompts.
	 The model supports Chromostereopsis, Denoising, Foveation, and Dynamic Range modification, and their permutations at different intensities using adjectives like \textit{``mildly''}, \textit{``slightly''}, and \textit{``lightly''} within a single inference, eliminating the need for daisy-chaining multiple models.
   Real-world input images in \textit{``foveate''} and \textit{``enhance dynamic range''} are attributed to Billy Wilson and Orgthingy. Remaining images are from our test set generated using Stable Diffusion \cite{rombach2022high}.
        }
\Description{Learned Multitasking Perceptual Graphics for Immersive Displays.}
\label{fig:teaser}
\end{teaserfigure}

%%
%% This command processes the author and affiliation and title
%% information and builds the first part of the formatted document.
\maketitle

\section{Introduction}
Immersive display technologies \cite{koulieris2019near}, including \AR glasses, \VR headsets, and large-format displays, are advancing towards more realistic image delivery.
\begin{wrapfigure}[28]{l}{0.23\columnwidth}
  \begin{center}
  \includegraphics[width=0.26\columnwidth]{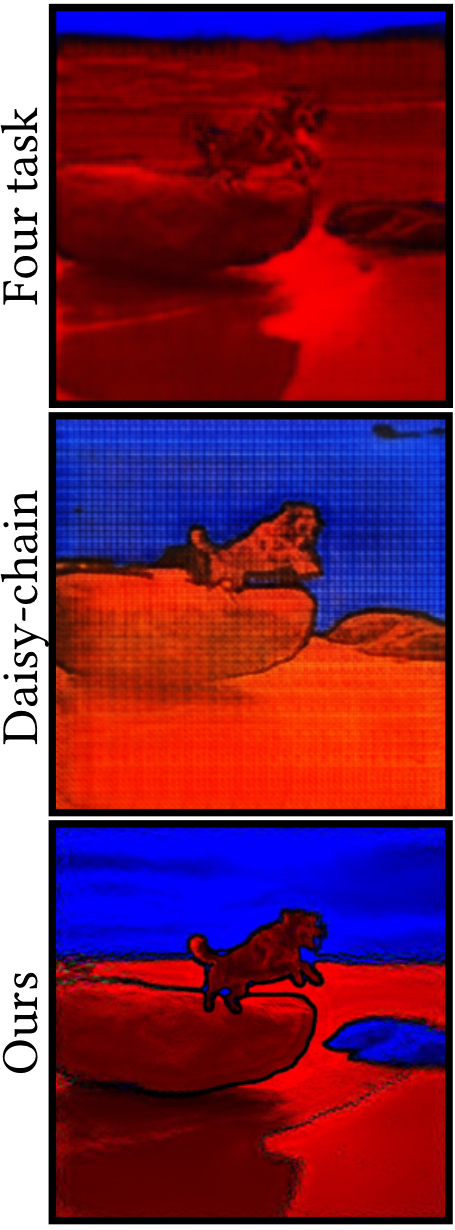}
  \end{center}
  \caption{Four perceptual tasks applied on various baseline models and ours. The capacities are equal per task, see \refSec{evaluation}.}
  \label{fig:practical_task_limits}
\end{wrapfigure}
However, these devices face constraints due to power, performance, and form-factor limitations, making on-device high-quality rendering a challenge.
Hence, researchers explore perceptual graphics methods such as foveated rendering \cite{tariq2024motion}, dynamic range enhancement \cite{marzuki2020perceptual}, image denoising \cite{conde2023perceptual}, and chromostereopsis \cite{westermann2022using} to enhance low quality images.
In practice, these emerging perceptual graphics methods need to be daisy-chained to produce images of high perceptual quality.
Daisy-chaining these perceptual models or learning models for desired combined perceptual tasks can quickly lead to poor image quality as shown in \refFig{practical_task_limits}.
Moreover, these perceptual effects use the same image attributes, such as depth, segmentation, and color, which creates redundancy in the model parameters and computation.
Mitigating this redundancy, a potentially more resource-efficient alternative to daisy-chaining is to combine multiple perceptual graphics methods with a multitask learning approach.
Also, recent works in generative models have demonstrated that combined learned multimodal approaches, enables a wide range of image-to-image translation tasks \cite{rombach2022high, huang2023composer}.
Inspired by these recent works, we propose to unify perceptual tasks in a single model to utilize bandwidth, and computational resources more efficiently while also supporting immersive displays in a device-agnostic manner, thereby meeting their unique rendering requirements.

Our work proposes a text-guided learned multitasking perceptual graphics model for immersive displays.
The input to this model is an RGB image and text prompt pair to guide the model to output perceptually enhanced images.
Our model is enabled by our new learned component, which we call as Embedding Mapper module.
This new module efficiently combines encoded RGB images and embeddings from text prompts at the bottleneck of a multitasking U-Net.
Leveraging multitask learning, our model supports various perceptual tasks and their combinations, including foveated rendering, dynamic range enhancement, image denoising, and chromostereopsis.
Utilizing text-embeddings rather than fixed-vectors benefits the practicality and flexibility of our model.
This choice enables adjustment of effect intensity and supports new tasks without requiring any changes in the architecture.
Furthermore, by maintaining a plain-text control interface, we ensure seamless compatibility with modern model-conditioning techniques.
To train our model, we introduce a new dataset comprising pairs of images and their text prompts, each representing a distinct perceptual effect.
Our model facilitates deployment on both desktop and embedded systems for immersive displays as it is lightweight and fast.
Furthermore, we validate the perceptual quality of the images generated by our model with a subjective experiment.
The source code of our learned model, along with our perceptual image dataset and model weights, can be found at \codebase.
Our contributions are as follows:
\begin{itemize}[topsep=0pt, itemsep=0pt, parsep=0pt, partopsep=0pt, leftmargin=*]
    \item \textit{Multitasking Perceptual Model.} Enabled by our new learned embedding mapper, which efficiently combines image and text embeddings, we propose a learned multitasking perceptual graphics model that transforms RGB images to various perceptually guided image styles.
    Our model can achieve hybrid tasks that are composed of novel permutations of individual tasks (\eg, enhance dynamic range and foveate) as well as controlling the degree of applied effect (\eg, mildly apply chromostereopsis) in a single inference step.
    Furthermore, we deploy our model on an NVIDIA Jetson Nano embedded device to demonstrate its effectiveness in computationally limited scenarios.
    \item \textit{Perceptual Evaluations.}
    We introduce a new dataset that contains pairs of images and their corresponding text prompts.
    Each pair represents an image-to-image translation of perceptual effects.
    We also provide a complete pipeline describing the image generation routine in our datasets.
    Utilizing this dataset and image generation pipelines, we validate the perceptual quality of the generated images from our model with a user study.
\end{itemize}

\section{Related Work}
Our work enables the simultaneous application of multiple perceptual graphics tasks to efficiently prepare media for immersive displays by leveraging multitask learning.
We review the relevant literature for each visual perception task we focus on, as well as for learned image processing methods and multitask learning approaches, to provide context for our contributions.

\begin{table}[!htb]
  \centering
  \caption{Overview of perceptual graphics techniques.
  Our work distinguishes itself by providing support for multiple perceptual effects in a single inference pass, with text-guidance and multitasking, while maintaining a lightweight architecture.
  Abbreviations: Foveation (F), Chromostereopsis (C), Image Denoising (ID), Dynamic Range Enhancement (DRE).}
  \label{tbl:method_comparison}
  \resizebox{\columnwidth}{!}{
    \begin{tabular}{lccccc}
    \hline
     & \textbf{Approach} & \textbf{\makecell{ Perceptual Tasks }} & \textbf{\makecell{ Text \\ Guidance }} & \textbf{\makecell{ Speed }} & \textbf{Multitasking} \\ \hline
    Walton \etal \cite{walton2021beyond}         &       Traditional       &  \medium{F}            & \low{None}                                                      & \high{\begin{tabular}{@{}c@{}}Real Time \end{tabular}} &  \low{None}                                                            \\ 
    Deza \etal \cite{deza2017towards}            &       Learned           &  \medium{F}            & \low{None}                                                      & \high{\begin{tabular}{@{}c@{}}Real Time \end{tabular}} &  \low{None}                                                            \\
    Westermann \etal \cite{westermann2022using}  &       Traditional       &  \medium{C}            & \low{None}                                                      & \high{\begin{tabular}{@{}c@{}}Real Time \end{tabular}} &  \low{None}                                                            \\ 
    Conde \etal \cite{conde2023perceptual}       &       Learned           &  \medium{ID}           & \low{None}                                                      & \high{\begin{tabular}{@{}c@{}}Real Time \end{tabular}} &  \low{None}                                                            \\ 
    Marzuki \etal \cite{marzuki2020perceptual}   &       Learned           &  \medium{DRE}          & \low{None}                                                      & \high{\begin{tabular}{@{}c@{}}Real Time \end{tabular}} &  \low{None}                                                            \\ 
    Afifi \etal \cite{afifi2020deep}             &       Learned           &  \low{None}            & \low{None}                                                      & \high{\begin{tabular}{@{}c@{}}Real Time \end{tabular}} & \medium{\begin{tabular}{@{}c@{}}Shared \\ Encoder\end{tabular}}        \\ 
    Sun \etal \cite{sun2021task}                 &       Learned           &  \low{None}            & \low{ \begin{tabular}{@{}c@{}}Fixed \\ Wording\end{tabular}}     & \high{\begin{tabular}{@{}c@{}}Real Time \end{tabular}} & \high{\begin{tabular}{@{}c@{}}Hard Parameter \\ Sharing\end{tabular}}  \\ 
    Huang \etal \cite{huang2023composer}         &       Learned           &  \low{None}            & \high{\begin{tabular}{@{}c@{}}Open \\ Ended\end{tabular}}       & \low{\begin{tabular}{@{}c@{}}Offline \end{tabular}}    & \high{\begin{tabular}{@{}c@{}}Hard Parameter \\ Sharing\end{tabular}}  \\
    Ours                                         &       Learned           &  \high{F, C, ID, DRE}  & \medium{\begin{tabular}{@{}c@{}}Semi-Open \\ Ended\end{tabular}} & \high{\begin{tabular}{@{}c@{}}Real Time \end{tabular}} & \high{\begin{tabular}{@{}c@{}}Hard Parameter \\ Sharing\end{tabular}} \\ 
    \hline
    \end{tabular}
    }
\end{table}

\subsection{Visual Perception Tasks}
Our work focuses on foveation, dynamic range enhancement, image denoising, and chromostereopsis.
Image denoising and dynamic range enhancement are well established tasks in the literature, whereas foveation and chromostereopsis tasks are actively being explored.
Here, we refer to dynamic range enhancement as increasing the bits used to represent brightness levels in an image.
Following the common literature in image denoising \cite{zhang2017beyond, ohayon2021high, conde2023perceptual} and dynamic range enhancement \cite{marzuki2020perceptual, choi2020task, yu2020low}, we train our model using pairs of images with low and high dynamic range and image pairs containing noisy and noise-free images, respectively.
We provide an overview of existing perceptual graphics techniques in \refTbl{method_comparison}.

\paragraph{Foveation}
Foveated rendering promises to reduce computational complexity by rendering perceptually accurate yet lower resolution images in the periphery, leveraging the variation in resolution acuity between the fovea and periphery in the \HVS.
Meng \etal \cite{10.1145/3203199} parameterizes foveated rendering by embedding polynomial kernel functions in the classic log-polar mapping, enabling variation in the sampling density and distribution of the rendered images. 
Another class of methods for foveated rendering uses metamers \cite{deza2017towards, walton2021beyond, tariq2024motion}, which are image patches that are perceptually indistinguishable despite being different in terms of pixel values.
Display hardware devices have recently adopted designs specifically catered towards foveated rendering \cite{kim2019foveated}, especially for \AR and \VR applications.
\textit{Our model follows the metamer approach proposed by Walton \etal \cite{walton2021beyond} to foveate images.}

\paragraph{Chromostereopsis}
Chromostereopsis \cite{bai2016perceived, 10.3389/fpsyg.2015.00337, westermann2022using} is a visual perceptual effect induced by using different colors in images, which leads to an illusion of perceived depth differences in various colors of the images.
Hong \etal \cite{hong2011depth} propose an algorithm to enhance perceived depth in images based on chromostereopsis and cubic effects.
Similarly, Jung \etal \cite{jung2012depth} introduce a depth map-based image enhancement algorithm utilizing chromostereopsis. 
Westermann \etal \cite{westermann2022using} recently proposed a novel rule-based method to enhance perceived depth in images, using results from a user study.
\textit{Building on Westermann \etal \cite{westermann2022using}, our work focuses on creating artistically appealing chromostereoptic images that maximize perceived depth.}

\subsection{Learned Multitasking Image Processing}

\paragraph{Learned image-to-image translation}
Isola \etal \cite{isola2017pix2pix} investigate conditional \GAN as a general-purpose solution for image-to-image translation tasks. 
Zhu \etal \cite{zhu2017unpaired} propose an unpaired image-to-image translation method using a cycle-consistent approach, which mitigates the need for paired training data. 
Additionally, Choi \etal \cite{Choi_2018_CVPR} propose a novel approach for multidomain image-to-image translations using a single model.
Recently, Ko \etal \cite{ko2023superstargan} introduced an independent classifier to enhance feature learning, addressing the limitations of Choi \etal's method \cite{Choi_2018_CVPR}.
Ke \etal \cite{ke2023neural} propose a memory efficient learned color mapping for color normalization and stylization.
Text-guided diffusion-based generative models have also been utilized for image-to-image translation \cite{brooks2023instructpix2pix, huang2023composer, lin2024pixwizard, Geng23instructdiff}.
However, these generative models are iterative, requiring multiple passes to obtain good quality images, making them unsuitable for interactive speeds.
For our perceptual tasks, we examined the diffusion-based approaches proposed by Brooks \etal \cite{brooks2023instructpix2pix} and InstructDiffusion \cite{Geng23instructdiff}.
Despite the inherent inference speed limitations, we included InstructDiffusion \cite{Geng23instructdiff} in our evaluation as a representative state-of-the-art diffusion-based method.
Additional results for Brooks \etal \cite{brooks2023instructpix2pix} can be found in the supplementary material (see \refSec{practical_experiments}).
\textit{Our work stands out as a lightweight application-specific solution suitable for embedded deployment, offering a text-guided perceptual graphics method for immersive displays.}

\paragraph{Multitask learning}
Introduced by Caruana \cite{caruana1997multitask}, \MTL is an inductive transfer mechanism aimed at improving generalization performance by leveraging the domain-specific information contained in the training signals of related tasks. 
In our work, we focus on hard parameter sharing, where all tasks share the parameters for the same model. 
The work by Sun \etal \cite{sun2020adashare} proposes an efficient sharing scheme that learns separate execution paths for different tasks.
In addition, Afifi \etal \cite{afifi2020deep} propose a deep multitask learning architecture for auto white balancing, utilizing a single encoder and multiple decoders, each corresponding to a specific task. 
Following up Afifi \etal \cite{afifi2020deep}, Sun \etal \cite{sun2021task} demonstrate multitasking with a single task-conditioned decoder.
Alternatively, diffusion-based generative models could be utilized in multitask learning scenarios \cite{huang2023composer}, but are not suitable for embedded development.
\textit{Similar to the architecture proposed by Sun \etal \cite{sun2021task}, our work utilizes text embeddings to learn multiple tasks with hard parameter sharing in the encoder and decoder. 
Our primary difference is in how we combine image and text embeddings using our embedding mapper.}

The aforementioned perceptual tasks and learned perceptual methods have been well explored individually.
Yet, the efficient unification of these tasks into a single model remains an open challenge.
\textit{Uniquely, our solution offers a text-guided multitasking model capable of applying all these perceptual tasks within a single, fast, cohesive model that can be deployed in embedded scenarios.}

\section{Text-Guided Perceptual Graphics}
\begin{figure}[!htb]
  \centering
  \includegraphics[width=\columnwidth]{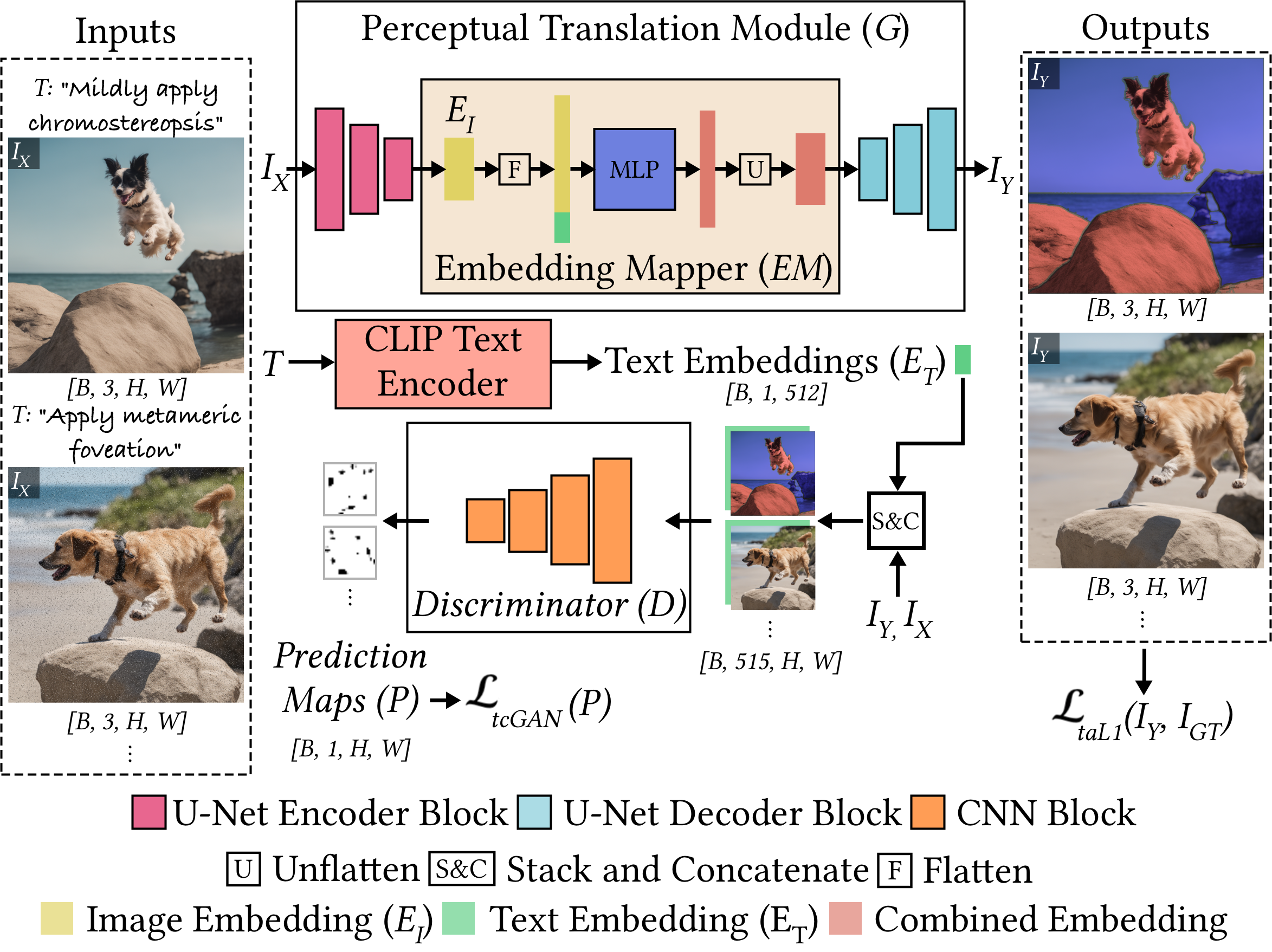}
  \caption{Our text-guided perceptual graphics model.
  Perceptual translation component ($G$) is conditioned on text embeddings ($E_T$) generated using $CLIP$ \cite{radford2021learning} and Embedding Mapper ($EM$).
  The $EM$ concatenates text embeddings ($E_T$) and image embeddings ($E_I$) to generate combined embedding ($E_C$). Task-aware discriminator ($D$) evaluates generated image ($I_Y$) for regularization.}
  \label{fig:architecture}
\end{figure}

Given an input RGB image and a text prompt describing the desired perceptual effect, our model in \refFig{architecture} applies the effect such as foveation, dynamic range enhancement, image denoising, and chromostereopsis, as well as their permutations at intended scales (\eg ``mildly,'' ``lightly'').

Our model comprises two main components: a perceptual translation component, $G$, and a task-aware discriminator component, $D$. 
Firstly, our perceptual translation component, $G$, a modified U-Net, transforms input images into perceptually enhanced output images.
This component incorporates our new Embedding Mapper module, $EM$, which conditions the perceptual translation based on embeddings derived from the provided input text prompts.
Secondly, our discriminator, $D$, guides the training of the perceptual translation component, $G$, by verifying the outputs according to the text embeddings.
During inference, $D$ is not deployed and is utilized solely during training to enhance the effectiveness of $G$.
Additionally, for extreme cases, including low bandwidth and low power, we utilize a pre-trained autoencoder \cite{rombach2022high} to stream the generated images, in a compressed format, to the user from a server with our model.

Our model is trained using a sample size adaptive loss function to scale the loss based on the number of samples available for each task and a task-aware adversarial loss to evaluate the generated images based on the task-specific text embeddings.
To train and evaluate our model, we introduce a **new** dataset comprising pairs of images, each representing a distinct perceptual enhancement.

\paragraph{Perceptual image translation component}
\label{sec:perceptual_translation}
Our first component, the perceptual image translation component $G$, has two primary objectives: (1) to enhance the visual perception of input images with desired effects, and (2) to ensure the model is lightweight and suitable for edge devices.
To meet all these requirements, we employ a U-Net architecture, similar to \cite{isola2017pix2pix}, with a modified bottleneck layer, and a pre-trained CLIP model \cite{radford2021learning} for text prompt guidance.
First, we transform the input text prompts, $T$, into text embeddings, $E_T \in \mathbb{R}^{B \times 1 \times 512}$, using the pre-trained CLIP model, $CLIP$.
In parallel, we encode the input images, $I_X \in \mathbb{R}^{B \times 3 \times H \times W}$, into image embeddings, $E_I \in \mathbb{R}^{B \times 1 \times 2048}$, using the encoder of the U-Net.
Following this, we flatten the text embeddings, $E_T$, and image embeddings, $E_I$, and concatenate them to form a single embedding, $E_{T+I}$.
This concatenated embedding, $E_{T+I}$, is then fed into our embedding mapper module, $EM$, consisting of an \MLP.
The $EM$ module maps the concatenated embedding, \(E_{T+I}\), to generate a combined embedding, $E_C \in \mathbb{R}^{B \times 1 \times 2048}$.
We also derived a simple text embedding generator that maps one-hot-encoded text embeddings to the $CLIP$ embedding space to test our model in isolation; see the supplementary \refSec{simple_task_embedding_generator} for details.
This operation not only merges text and image information into a single embedding but also ensures that the dimensionality of the resulting combined embedding, $E_C$, is compatible with the symmetric encoder-decoder U-Net architecture.
The combined embedding, $E_C$, is then unflattened and fed to the bottleneck of our U-Net decoder to produce the perceptually enhanced image, $I_Y \in \mathbb{R}^{B \times 3 \times H \times W}$.
\textit{Our $EM$ offers a unique application-specific conditioning solution without requiring computationally demanding conditioning at every layer of a U-Net \cite{karras2019style,patashnik2021styleclip} or a dedicated network for merging images and texts in the input \cite{xia2021tedigan} or enlarged decoder capacity due to size mismatch originated from concatenating text and image embeddings at the bottleneck \cite{dong2017semantic,vo2022paired}.}
The architecture of our perceptual translation module, $G$, and our strategy for guiding the model with text prompts are illustrated in \refFig{architecture}.
Detailed configurations of $EM$ and the pre-trained CLIP model are in the supplementary's \refSec{embedding_mapper_details}.

\paragraph{Sample size adaptive loss}
When introducing new tasks in the training of $G$, the number of samples available for a new task may be different than the existing tasks.
Thus, we introduce a sample count adaptive loss function that regularizes $G$ in training according to the number of samples available for each task.
Considering the largest sample count among all tasks, $SC_{MAX}$, and the sample counts for each task, $SC_T$, we calculate boosting factors, $B_T$, inversely proportional to the sample counts, capped by a maximum boost coefficient, $B_{MAX}$.
These boosting factors are then used to scale the L1 loss for each task,
\begin{equation}
\label{eq:sample_size_adaptive_loss}
  \begin{split}
    B_T &= 1 + (B_{MAX} (1 - \frac{SC_T}{SC_{MAX}})), \\
    \mathcal{L}_{\text{taL1}} &=  \mathcal{L}_{L1}(I_Y, I_{GT}) B_T.
  \end{split}
\end{equation}
By amplifying the loss inversely proportional to the sample counts for tasks with fewer samples, we ensure optimal use of available data and encourage the optimization process to allocate greater updates for these underrepresented tasks.
We further evaluate sample size adaptive L1 loss in our ablation study in \refSec{evaluation}.

\paragraph{Task-aware discriminator component}
\label{sec:multitasking_discriminator}
It has been demonstrated that utilizing conditional \GAN loss \cite{isola2017pix2pix} effectively regularizes image translation tasks by improving the quality of generated images while preserving the original content.
Building on the image-based conditioning proposed by Isola \etal \cite{isola2017pix2pix}, we extend this approach to include task conditioning in our adversarial loss.
To guide the training of our model, we employ a multitasking task-aware discriminator, $D$.
This discriminator processes the generated image, $I_Y$, the input image, $I_X$, and the task-specific text embeddings, $E_T$, to generate probability maps that facilitates the calculation of the task-aware adversarial loss.
To support this operation, the text embeddings, $E_T$, are stacked and concatenated along the channel dimension of the $I_Y$ and $I_X$.
The resulting tensor, which comprises both the image and task information, is subsequently fed into $D$ to obtain prediction maps of $P_0 \in \mathbb{R}^{B \times 1 \times H \times W}$ and $P_1 \in \mathbb{R}^{B \times 1 \times H \times W}$.
Our multitasking task-aware discriminator is illustrated in \refFig{architecture}.
We leverage $P_0$ and $P_1$ to provide a pixel-wise estimation of the likelihood that each pixel belongs to the perceptually enhanced image, $I_Y$.
$P_0$ and $P_1$ generated by $D$ are used to enhance task-aware guidance during the training of $G$ as shown in the following equation:
\begin{equation}
\label{eq:task_aware_cgan_loss}
  \begin{split}
    P_0 &= D(I_X , I_{GT}, E_T), \quad P_1 = D(I_X , I_Y, E_T), \\
    \mathcal{L}_{\text{tcGAN}} &= \mathbb{E}_{I_X, I_{GT}, E_T} [\log P_0] + \mathbb{E}_{I_X, I_Y, E_T} [\log(1 - P_1)].
  \end{split}
\end{equation}

\paragraph{Objective functions and training procedure}
\label{sec:objective_functions}
We guide the training of our model by utilizing the following functions: (1) a sample size adaptive L1 loss, and (2) a task-aware adversarial loss.
Our total loss function is formulated as shown in \refEq{total_loss}, where $\lambda$ is a hyperparameter that adjusts the contribution of the sample size adaptive loss to the total loss,
\begin{equation}
  \mathcal{L}_{\text{total}} = \mathcal{L}_{\text{tcGAN}} + \lambda \mathcal{L}_{\text{taL1}}.
\label{eq:total_loss}
\end{equation}
We use a two-phase training strategy.
Initially, our training dataset is restricted to image pairs from single tasks, which allows the model to focus on learning each task independently.
After completing this phase, we expand our training dataset by incorporating both single and combined task image pairs, and continue the training process.
Hyperparameters and training details are available in the supplementary's \refSec{hyperparameters_training_details}.

\begin{wrapfigure}[23]{l}{0.24\columnwidth}
  \begin{center}
    \includegraphics[width=0.28\columnwidth]{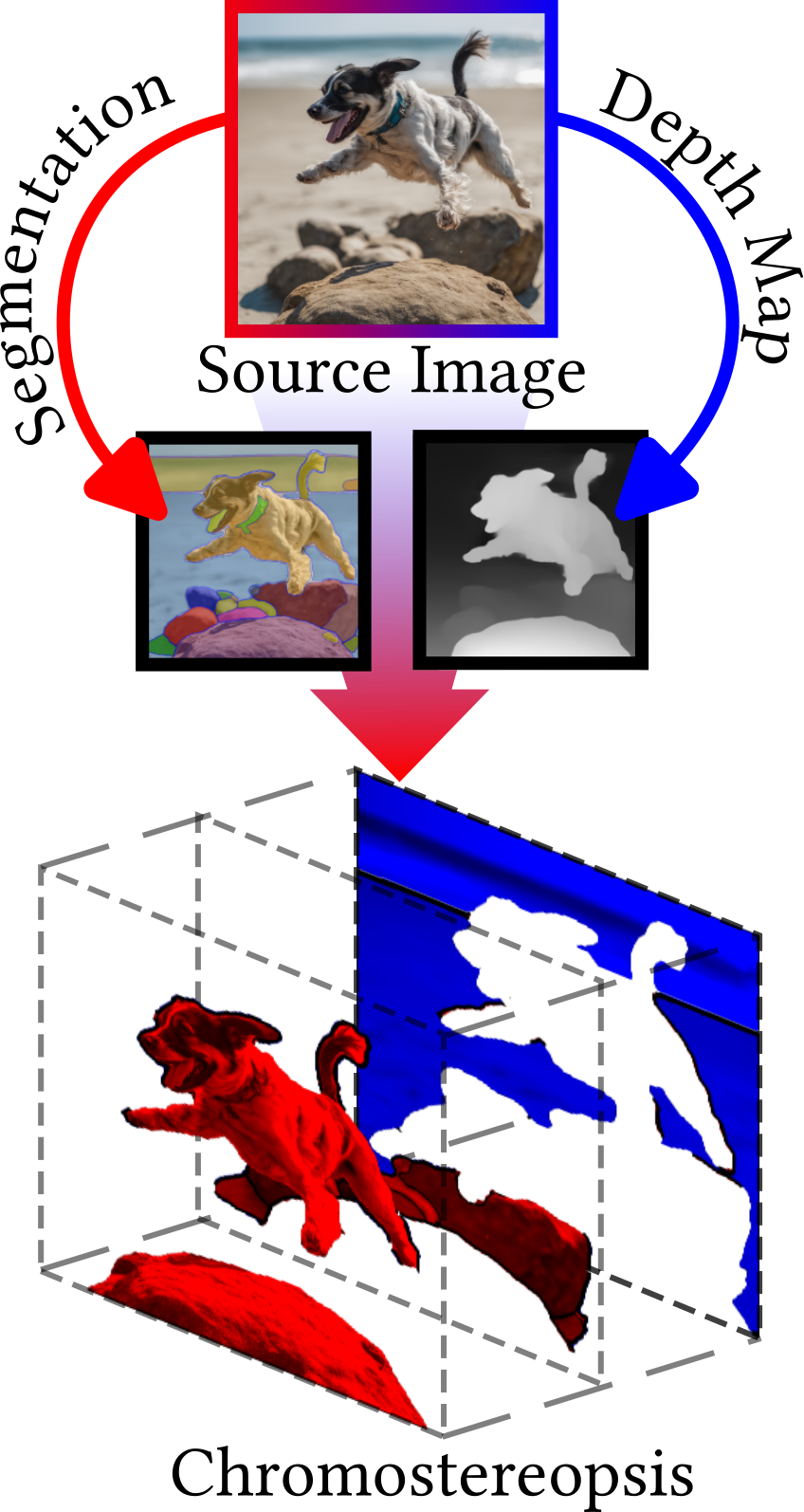}
  \end{center}
  \caption{Utilizing depth \cite{ranftl2021vision} and segmentation \cite{kirillov2023segment} estimation to generate chromostereopsis images.}
  \label{fig:chromo_methodology}
\end{wrapfigure}

\paragraph{Embedding streaming component}
\label{sec:embedding_streaming}
Optionally, in scenarios where computational resources and bandwidth are extremely limited, streaming the generated images in a compressed form to the user can be beneficial for presenting high quality media.
For such cases, our model can be deployed on a more powerful server, where the generated images are compressed and streamed to the client.
To support this, we utilize a distilled version of the pre-trained encoder and decoder \cite{rombach2022high}, where images are compressed using the encoder.
The compressed images are then streamed to the client, where they are decompressed by the decoder to be displayed.
This component is optional and can be disabled in scenarios where computational and bandwidth resources are not a concern.
Practical details for our streaming component are available in the supplementary's \refSec{embedding_streaming_component_details}.

\paragraph{Perceptual graphics dataset}
As our model learns a mapping from observed images, $I_X$, and text embeddings, $E_T$, to perceptually enhanced images, $I_Y$, denoted as $G : (I_X, E_T) \rightarrow I_Y$, it requires a dedicated set of text prompts paired with corresponding images for training.
Generating such paired data can be particularly challenging for complex tasks like chromostereopsis, and the challenge can easily stack up as in our combined task cases.
Thus, we propose a dataset containing 8800 image pairs with corresponding text prompts, each at a resolution of 1024x1024 pixels, distributed equally across various perceptual tasks, including foveated rendering, dynamic range enhancement, image denoising, chromostereopsis, and their permutations.
Additionally, the prompts in our dataset feature adjectives such as “mildly,” “lightly,” and “slightly” to control the intensity of the applied effect.
For all of our tasks, we generate RGB source images using Stable Diffusion \cite{rombach2022high}.
Our foveated image examples rely on Walton \etal~\shortcite{walton2021beyond}, for dynamic range enhancement we clip the dynamic range of the generated images from 8 bits to 4 bits, and for image denoising we add salt and pepper noise to the generated images.
To induce the chromostereopsis effect, we first generate a depth map from the ground truth images using the monocular depth estimation method by Ranftl \etal \cite{ranftl2021vision}, then segment these images following the method by Kirillov \etal \cite{kirillov2023segment}.
The final chromostereopsis images are produced by adjusting the hue of the foreground segments to red and the background segments to blue, based on the average depth of each segment, as proposed by Westermann \etal \cite{westermann2022using} as shown in \refFig{chromo_methodology}.
For the combined task image pairs, we apply the related methodology of each individual task consecutively.
All supported tasks are listed in \refTbl{quantitative_results}, and details about the datasets, including the supported adjectives, can be found in the supplementary \refSec{dataset_samples}.

\section{Evaluation}
\label{sec:evaluation}
\begin{figure*}[!htb]
  \centering
   \includegraphics[width=\textwidth]{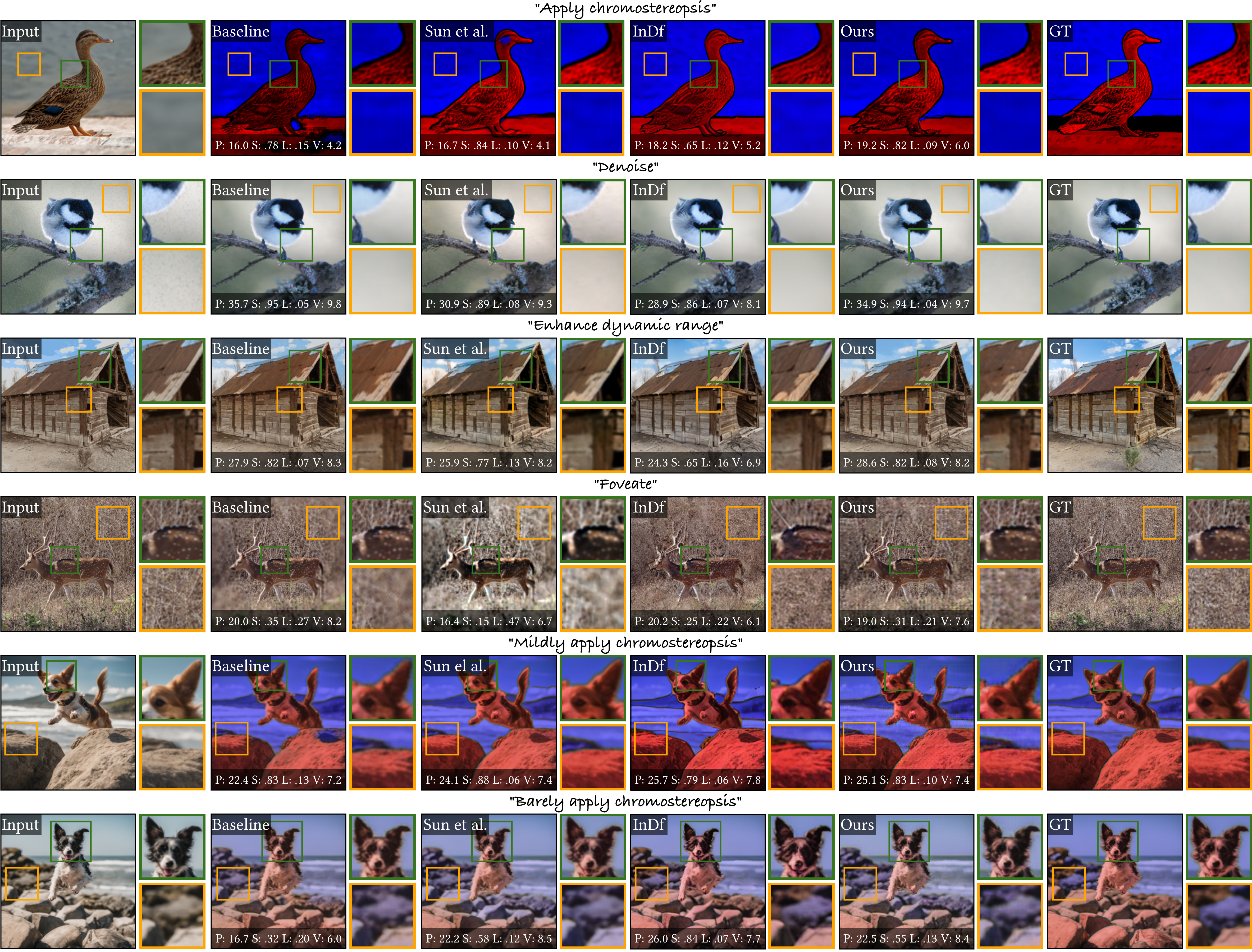}
   \caption{
   Results of our multitasking perceptual model compared to Sun \etal \cite{sun2021task}, InstructDiffusion (InDf) \cite{Geng23instructdiff}, and task-specific Vanilla U-Nets (Baseline).
   Metrics: \PSNR (P) $\uparrow$ \SSIM (S) $\uparrow$ \LPIPS (L) $\downarrow$ \FovVideoVDP (V) $\uparrow$.
   Dynamic range enhancement results are best seen on video displays.
   (Real-world images in rows 1-4 attributed to Miguel Discart, Michael Kuhn, James Marvin Phelps, and Lahar Jadav. Remaining images from our test set generated using Stable Diffusion \cite{rombach2022high}.)}
   \label{fig:results0}
\end{figure*}

We evaluate our learned model in terms of image quality (see \refTbl{quantitative_results}) and inference speed (see \refTbl{performance_results}).
To assess image quality, we employ metrics such as \PSNR, \SSIM, \LPIPS \cite{zhang2018unreasonable}, and \FovVideoVDP \cite{mantiuk2021fovvideovdp}.
We conducted a user study to further confirm that the image quality produced by our model is comparable to the state-of-the-art methods.
Visual results for both synthetic and real-world images are presented in \refFig{results0}.
For more visual results beyond \refFig{results0}, consult our supplementary's \refSec{additional_visual_results}.

\begin{wrapfigure}[23]{l}{0.24\columnwidth}
  \begin{center}
  \includegraphics[width=0.25\columnwidth]{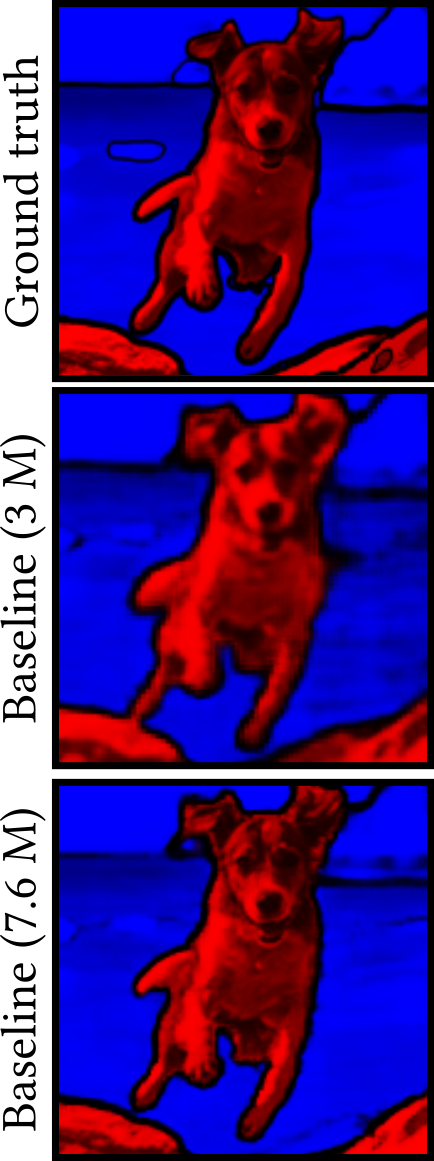}
  \end{center}
  \caption{Lower capacity per task leads to visual distortions in our baseline models.}
  \label{fig:lower_capacity}
\end{wrapfigure}

\paragraph{Inference speed}
We compare our model's performance against Sun \etal \cite{sun2021task}, InstructDiffusion \cite{Geng23instructdiff}, and relevant baseline models to assess its image quality and inference speed using 32-bit precision (FP32).
Note that the method proposed by Sun \etal \cite{sun2021task} does not support text guidance and is not fully equivalent to our proposed method in this respect.
To extend the evaluation further, we formulated several baseline models, which consists of vanilla U-Net models trained on our dataset and deployed in three different settings.
These settings are single-task, daisy-chain, and N-task, where N represents the number of simultaneously applied tasks.
Here, a single-task model refers to a model trained for a specific task (e.g., foveation); a daisy-chain model refers to running single-task models consecutively (e.g., image denoising and foveation); and an N-task model refers to a model trained with a specific combination of tasks to perform all tasks in a single inference.
For a fair comparison, we set the model capacity of all models to be equal per task (\ie $\sim7.6$ million parameters per task) as reported in \refTbl{performance_results}.
\begin{table}[!t]
  \centering
  \caption{
    Performance evaluation results of our model, Sun \etal \cite{sun2021task}, InstructDiffusion \cite{Geng23instructdiff}, and baselines (vanilla U-Net) on desktop (NVIDIA RTX 3090) and embedded devices (NVIDIA Jetson Nano) using 32-bit precision (FP32).
	}
  \label{tbl:performance_results}
  \resizebox{\columnwidth}{!}{
  \begin{tabular}{llccc}
    \hline
    \textbf{Device} & \textbf{Model} & \textbf{\makecell{ Inference \\ Speed (ms) }} & \textbf{\makecell{ Parameter \\ Count (M) }} & \textbf{\makecell{ Task \\ Count }} \\ \hline
    \rowcolor{table_shade2}
    \cellcolor{table_shade1}                           & Single-task & 129.56 ms  & 7.656 M & 1 \\
    \rowcolor{table_shade2}
    \cellcolor{table_shade1}                           & Daisy-chain & 409.16 ms  & 15.312 M & 2 \\
    \rowcolor{table_shade2}
    \cellcolor{table_shade1}                           & Daisy-chain & 810.28 ms  & 30.624 M & 4 \\
    \rowcolor{table_shade2}
    \cellcolor{table_shade1}                           & Two-task & 129.56 ms  & 7.656 M & 2 \\
    \rowcolor{table_shade2}
    \cellcolor{table_shade1}                           & Four-task & 129.56 ms  & 7.656 M & 4 \\
    \rowcolor{table_shade1}
    \cellcolor{table_shade1}                           & \textbf{Ours (streamed)} & 179.14 ms  & 1.222 M & 1-4 \\
    \rowcolor{table_shade1}                                  
    \multirow{-7}{*}{\cellcolor{table_shade1}Embedded} & \textbf{Ours} & 260.82 ms &  50.593 M & 1-4 \\
    \rowcolor{table_shade4} 
    \cellcolor{table_shade3}                           & Single-task & 1.34 ms & 7.656 M & 1 \\
    \rowcolor{table_shade4} 
    \cellcolor{table_shade3}                           & Daisy-chain & 3.79 ms & 15.312 M & 2 \\
    \rowcolor{table_shade4} 
    \cellcolor{table_shade3}                           & Daisy-chain & 7.65 ms & 30.624 M & 4 \\
    \rowcolor{table_shade4} 
    \cellcolor{table_shade3}                           & Two-task & 1.34 ms & 7.656 M & 2 \\
    \rowcolor{table_shade4} 
    \cellcolor{table_shade3}                           & Four-task & 1.34 ms & 7.656 M & 4 \\
    \rowcolor{table_shade4} 
    \cellcolor{table_shade3}                           & Sun \etal \cite{sun2021task} & 4.81 ms & 22.124 M & 1-4 \\
    \rowcolor{table_shade4}
    \cellcolor{table_shade3}                           & InstructDiffusion \cite{Geng23instructdiff} & 536.96 ms & 859.530 M & 1-4 \\
    \rowcolor{table_shade3}
    \multirow{-8}{*}{\cellcolor{table_shade3}Desktop}  & \textbf{Ours} & 1.74 ms & 50.593 M & 1-4 \\ \hline
    \end{tabular}
    }
\end{table}
Although our model has $\sim50$ million parameters excluding $D$; the parameters for the U-Net for our model consumes only $\sim3$ million parameters, which is half the size of other baseline models with $\sim7.6$ million parameters dedicated to the U-Net at minimum.
The remaining $\sim47$ million parameters are used for the Embedding Mapper module ($EM$), where the feature sizes are close to bottleneck feature size in our U-Net.
\textit{Dedicating more parameters to the $EM$ with small feature sizes help us to achieve inference speeds as fast as a single-task model while supporting all the benefits of text-guidance and multitasking with a single model.}
If the baseline models use lower capacity in their U-Net following our model ($\sim3$ million parameters), they render visually distorted blurry images as sampled in \refFig{lower_capacity}, making these lower capacity baseline models unusable for comparison.
Consult our supplementary's \refSec{model_architecture_details} for the lower capacity U-Net details.
Our baselines (single-task, daisy-chain, and N-task) and the work by Sun \etal \cite{sun2021task} are also limited in the number of supported tasks and do not offer full flexibility to blend tasks at will --lightly foveate and fully denoise is not an option for a foveation and denosing baseline--.
This necessitates the training of many models for many tasks, making model management an issue especially for embedded devices.
In comparison to the single-task and two-task models, \refTbl{performance_results} shows that our model has similar inference speeds.
Daisy-chain models are naturally slower than our model in inference speed due to dedicating larger capacities.
In contrast to our method, InstructDiffusion \cite{Geng23instructdiff} has significantly slower inference due to its larger model size and iterative diffusion process.
As observed in \refTbl{performance_results}, among flexible and controllable models, ours is the fastest, achieving inference speeds 2.5 times faster than the nearest comparable approach.

\paragraph{Quantitative image quality}
We evaluate the visual quality of our model, method proposed by Sun \etal \cite{sun2021task}, InstructDiffusion \cite{Geng23instructdiff}, and the baselines with established metrics in \refTbl{quantitative_results} and provide \refFig{results0} to sample their task performance qualitatively for both synthetic and real-world images.
Results in \refTbl{quantitative_results} are averaged over the test split.
\begin{table}[!t]
  \centering
  \caption{
    Quantitative image quality of our model, Sun \etal \cite{sun2021task}, InstructDiffusion \cite{Geng23instructdiff}, and baseline models (vanilla U-Nets).
    Abbreviations: Foveation (F), Chromostereopsis (C), Image Denoising (ID), Dynamic Range Enhancement (DRE).
  }
  \label{tbl:quantitative_results}
  \resizebox{\columnwidth}{!}{
    \begin{tabular}{llcccc}
      \hline
      \textbf{Task} & \textbf{Model} & \textbf{PSNR (dB) $\uparrow$} & \textbf{SSIM $\uparrow$} & \textbf{LPIPS $\downarrow$} & \textbf{FovVideoVDP $\uparrow$}\\ \hline
    \rowcolor{table_shade2}
    \cellcolor{table_shade1}                                                                          & Single-task & \textbf{27.43} & \textbf{0.79} & 0.18 & \textbf{9.23} \\
    \rowcolor{table_shade2}
    \cellcolor{table_shade1}                                                                          & Sun \etal \cite{sun2021task} & 27.12 & 0.78 & 0.23 & 9.17 \\
    \rowcolor{table_shade2}
    \cellcolor{table_shade1}                                                                          & InstructDiffusion \cite{Geng23instructdiff} & 25.74 & 0.75 & 0.14 & 8.92 \\
    \rowcolor{table_shade1}
    \multirow{-4}{*}{\cellcolor{table_shade1}F}                                               & \textbf{Ours} & 25.64 \textcolor{red}{(-1.79)} & 0.74 \textcolor{red}{(-0.05)} & \textbf{0.10} \textcolor{Green}{(+0.04)} & 9.01 \textcolor{red}{(-0.22)} \\
    \rowcolor{table_shade4}
    \cellcolor{table_shade3}                                                                          & Single-task & \textbf{33.38} & \textbf{0.92} & 0.05 & \textbf{9.25} \\
    \rowcolor{table_shade4}
    \cellcolor{table_shade3}                                                                          & Sun \etal \cite{sun2021task} & 31.92 & 0.91 & 0.06 & 9.21 \\
    \rowcolor{table_shade4}
    \cellcolor{table_shade3}                                                                          & InstructDiffusion \cite{Geng23instructdiff} & 29.89 & 0.90 & \textbf{0.03} & 9.19 \\
    \rowcolor{table_shade3}
    \multirow{-4}{*}{\cellcolor{table_shade3}DRE}            & \textbf{Ours} & 31.07 \textcolor{red}{(-2.31)} & 0.88 \textcolor{red}{(-0.04)} & 0.08 \textcolor{red}{(-0.03)} & 9.06 \textcolor{red}{(-0.19)} \\ 
    \rowcolor{table_shade2}
    \cellcolor{table_shade1}                                                                          & Single-task & \textbf{35.90} & \textbf{0.95} & \textbf{0.03} & \textbf{9.79} \\
    \rowcolor{table_shade2}
    \cellcolor{table_shade1}                                                                          & Sun \etal \cite{sun2021task} & 34.26 & 0.94 & 0.04 & 9.76 \\
    \rowcolor{table_shade2}
    \cellcolor{table_shade1}                                                                          & InstructDiffusion \cite{Geng23instructdiff} & 29.49 & 0.83 & 0.07 & 9.53 \\
    \rowcolor{table_shade1} 
    \multirow{-4}{*}{\cellcolor{table_shade1}ID}                      & \textbf{Ours} & 34.05 \textcolor{red}{(-1.85)} & 0.92 \textcolor{red}{(-0.03)} & 0.08 \textcolor{red}{(-0.05)} & 9.77 \textcolor{red}{(-0.02)} \\
    \rowcolor{table_shade4}
    \cellcolor{table_shade3}                                                                          & Single-task & 16.87 & 0.81 & 0.14 & 5.53 \\
    \rowcolor{table_shade4}
    \cellcolor{table_shade3}                                                                          & Sun \etal \cite{sun2021task} & 17.02 & 0.80 & 0.14 & \textbf{5.80} \\
    \rowcolor{table_shade4}
    \cellcolor{table_shade3}                                                                          & InstructDiffusion \cite{Geng23instructdiff} & 17.03 & 0.75 & 0.13 & 5.92 \\
    \rowcolor{table_shade3}
    \multirow{-4}{*}{\cellcolor{table_shade3}C}                                        & \textbf{Ours} & \textbf{17.04} \textcolor{Green}{(+0.01)} & \textbf{0.81} (0.00) & \textbf{0.13} (0.00) & 5.54 \textcolor{red}{(-0.38)} \\ 
    \rowcolor{table_shade2}
    \cellcolor{table_shade1}                                                                      & Two-task & 16.94 & \textbf{0.81} & 0.15 & 5.45 \\
    \rowcolor{table_shade2}
    \cellcolor{table_shade1}                                                                          & Daisy-chain & 16.02 & 0.73 & 0.16 & 5.43 \\
    \rowcolor{table_shade2}
    \cellcolor{table_shade1}                                                                          & Sun \etal \cite{sun2021task} & \textbf{17.80} & \textbf{0.81} & \textbf{0.11} & 5.44 \\
    \rowcolor{table_shade2}
    \cellcolor{table_shade1}                                                                          & InstructDiffusion \cite{Geng23instructdiff} & 17.43 & 0.67 & 0.12 & 4.44 \\
    \rowcolor{table_shade1} 
    \multirow{-5}{*}{\cellcolor{table_shade1}ID and C} & \textbf{Ours} & 16.74 \textcolor{red}{(-1.06)} & 0.80 \textcolor{red}{(-0.01)} & 0.14 \textcolor{red}{(-0.03)} & \textbf{5.47} \textcolor{Green}{(+0.02)}\\
    \rowcolor{table_shade4}
    \cellcolor{table_shade3}                                                                           & Two-task & 16.49 & \textbf{0.80} & 0.14 & 5.29 \\
    \rowcolor{table_shade4}
    \cellcolor{table_shade3}                                                                          & Daisy-chain & 16.27 & 0.80 & 0.14 & 5.27 \\
    \rowcolor{table_shade4}
    \cellcolor{table_shade3}                                                                          & Sun \etal \cite{sun2021task} & 16.53 & \textbf{0.80} & 0.15 & 5.33 \\
    \rowcolor{table_shade4}
    \cellcolor{table_shade3}                                                                          & InstructDiffusion \cite{Geng23instructdiff} & \textbf{17.61} & 0.74 & \textbf{0.12} & 5.32 \\
    \rowcolor{table_shade3}
    \multirow{-5}{*}{\cellcolor{table_shade3}DRE and C} & \textbf{Ours} & 15.91 \textcolor{red}{(-1.70)} & 0.78 \textcolor{red}{(-0.02)} & 0.16 \textcolor{red}{(-0.04)} & \textbf{5.36} \textcolor{Green}{(+0.03)} \\ 
    \rowcolor{table_shade2} 
    \cellcolor{table_shade1}                                                                            & Two-task & \textbf{27.15} & \textbf{0.78} & 0.22 & 9.16 \\
    \rowcolor{table_shade2}
    \cellcolor{table_shade1}                                                                          & Daisy-chain & \textbf{27.15} & \textbf{0.78} & 0.20 & \textbf{9.17} \\
    \rowcolor{table_shade2}
    \cellcolor{table_shade1}                                                                          & Sun \etal \cite{sun2021task} & 26.98 & 0.77 & 0.23 & 9.14 \\
    \rowcolor{table_shade2}
    \cellcolor{table_shade1}                                                                          & InstructDiffusion \cite{Geng23instructdiff} & 25.16 & 0.71 & 0.19 & 8.87 \\
    \rowcolor{table_shade1} 
    \multirow{-5}{*}{\cellcolor{table_shade1}ID and F} & \textbf{Ours} & 25.65 \textcolor{red}{(-1.50)} & 0.71 \textcolor{red}{(-0.07)} & \textbf{0.11} \textcolor{Green}{(+0.8)} & 8.98 \textcolor{red}{(-0.19)} \\
    \rowcolor{table_shade4}
    \cellcolor{table_shade3}                                                                              & Two-task &  \textbf{26.60} & 0.75 & 0.23 & \textbf{8.85} \\
    \rowcolor{table_shade4}
    \cellcolor{table_shade3}                                                                          & Daisy-chain & 26.59 & \textbf{0.76} & 0.21 & 8.84 \\
    \rowcolor{table_shade4}
    \cellcolor{table_shade3}                                                                          & Sun \etal \cite{sun2021task} & 26.35 & 0.74 & 0.25 & 8.77 \\
    \rowcolor{table_shade4}
    \cellcolor{table_shade3}                                                                          & InstructDiffusion \cite{Geng23instructdiff} & 24.65 & 0.73 & 0.12 & 8.58 \\
    \rowcolor{table_shade3}
    \multirow{-5}{*}{\cellcolor{table_shade3}DRE and F} & \textbf{Ours} & 25.06 \textcolor{red}{(-1.54)} & 0.69 \textcolor{red}{(-0.07)} & \textbf{0.11} \textcolor{Green}{(+0.01)} & 8.58 \textcolor{red}{(-0.27)} \\ 
    \rowcolor{table_shade2}
    \cellcolor{table_shade1}                                                                              & Four-task & 16.27 & 0.62 & 0.22 & 5.30 \\
    \rowcolor{table_shade2}
    \cellcolor{table_shade1}                                                                          & Daisy-chain & 12.46 & 0.30 & 0.36 & 4.06 \\
    \rowcolor{table_shade2}
    \cellcolor{table_shade1}                                                                          & Sun \etal \cite{sun2021task} & 17.05 & 0.62 & 0.18 & 5.46 \\
    \rowcolor{table_shade2}
    \cellcolor{table_shade1}                                                                          & InstructDiffusion \cite{Geng23instructdiff} & 17.02 & 0.57 & 0.14 & 5.53 \\
    \rowcolor{table_shade1}
    \multirow{-5}{*}{\cellcolor{table_shade1}\makecell[l]{DRE and ID \\ and F and C}} & \textbf{Ours} & \textbf{17.14} \textcolor{Green}{(+0.09)} &  \textbf{0.66} \textcolor{Green}{(+0.04)} & \textbf{0.14} \textcolor{Green}{(+0.04)} & \textbf{5.61} \textcolor{Green}{(+0.08)} \\ \hline
    \end{tabular}
    }
\end{table}
Across a variety of metrics, our models achieve on-par performance in terms of image quality when compared to competitor models.
Beyond image quality, our approach offers notable advantages in versatility, task adaptability, precise control over effect intensity, and the flexibility to deploy either fully or partially on an embedded device.
By using text-guidance adjectives (\eg, “strongly foveate”) to specify different effect intensities, our model can dynamically adjust perceptual effects, as demonstrated in \refFig{results0}.
In contrast, while baseline models lack on-the-fly adaptability, Sun \etal \cite{sun2021task} offer similar functionality with inference speed 2.5 times slower.
InstructDiffusion \cite{Geng23instructdiff} offers on-the-fly adaptability; however, its high computational demands make it unsuitable for real-time applications and deployment on embedded devices.
Additional visual results for other supported tasks are provided in our supplementary’s \refSec{additional_visual_results}.
Video results of our method are also included, with further details available in \refSec{video_inference_details}.

\paragraph{Supporting complex tasks}
\label{sec:supporting_complex_tasks}
The increasing number of tasks in a combined task compounds the overall complexity, making it more challenging to produce high-quality images.
As indicated in the last row of \refTbl{quantitative_results} and in \refFig{practical_task_limits}, when the task count increases to four, our model surpasses both the daisy-chain method and the four-task models across all measured metrics.
This indicates that the ability of the four-task model becomes insufficient for generating high-quality images.
In contrast, our multitasking approach provides a flexible solution, allowing for the blending of tasks and the generation of high-quality images using a single model.

\begin{wrapfigure}[24]{l}{0.22\columnwidth}
  \begin{center}
  \includegraphics[width=0.22\columnwidth]{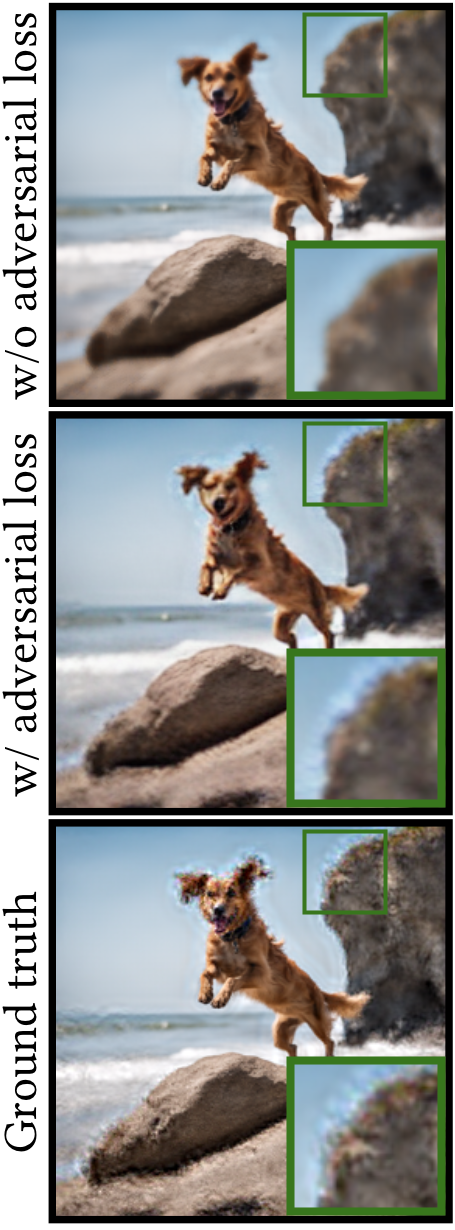}
  \end{center}
  \caption{Our task-aware adversarial loss preserve features at the periphery in foveated rendering.}
  \label{fig:task_aware_adversarial_loss_abl}
\end{wrapfigure}

\paragraph{Ablation study}
\begin{figure}[!t]
  \centering
  \includegraphics[width=\columnwidth]{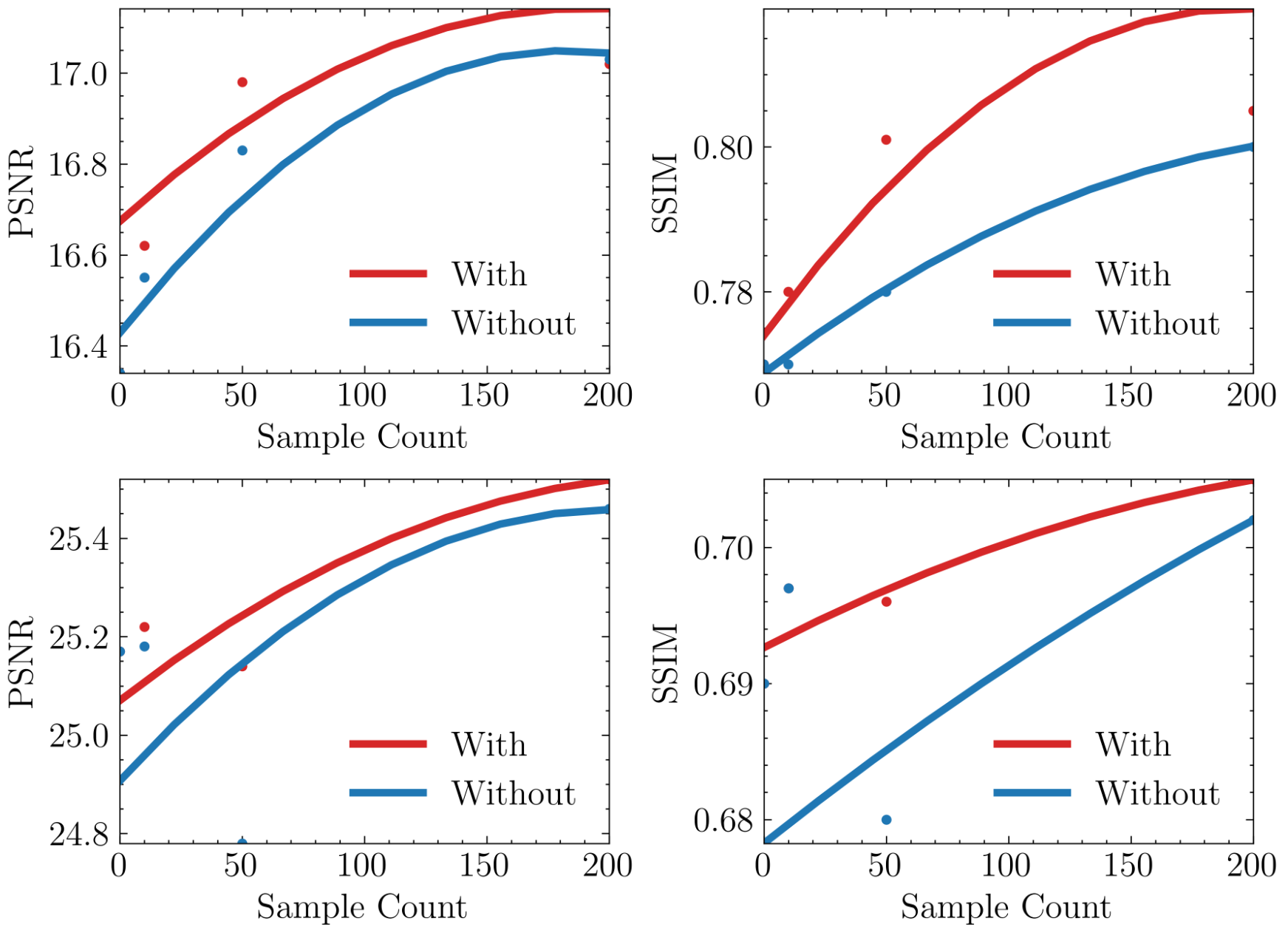}
  \caption{Performance of our model in terms of \PSNR and \SSIM across various training sample sizes for two reduced tasks: image denoising and chromostereopsis (top row) and denoise and foveate (bottom row).
  \textcolor{red}{Red} curves represent results with sample size adaptive loss, while \textcolor{blue}{blue} curves represent results without it.
  The plotted curves are second-degree polynomials fitted to the data.
  }
  \label{fig:sample_size_adaptive_loss_abl}
\end{figure}
We conducted ablation studies to validate the effectiveness of our proposed components.
Specifically, we evaluated the performance contribution of each component of our loss function.
Additionally, we inspected how the parameter count allocated to the $EM$ impacts the model's performance.
\textbf{Task-aware adversarial loss.} \refFig{task_aware_adversarial_loss_abl} presents a representative sample image that demonstrates the impact of the task-aware adversarial loss.
When we include the task-aware adversarial loss in our model, we observe that the model preserves high-frequency details of the image.
Without the task-aware adversarial loss, the model fails to preserve these details similar to the baseline models.
We invite readers to observe the high-frequency details in the foveated regions of the images in \refFig{task_aware_adversarial_loss_abl}.
\textbf{Sample size adaptive loss.} 
In our training dataset, we reduced the sample count of a set of tasks to simulate a low sample count scenario.
The reduced sample sizes are as follows: 0, 50, 100, 150, and 200, whereas a non-reduced task contain 880 samples.
We measured the performance using \PSNR and \SSIM (\refFig{sample_size_adaptive_loss_abl}), showing that the sample size adaptive loss improves performance on low-sample tasks.
\textbf{EM parameter count.} 
Beginning with the smallest possible $EM$ ($\sim50$M parameters), constrained by image dimensions, we incrementally increased its capacity to $\sim60$M and $\sim100$M.
Our observations indicate that increasing the capacity of $EM$ does not affect the model's performance.
Sample images generated using $EM$ of different sizes, along with their corresponding performance metrics, can be found in the supplementary's \refSec{embedding_mapper_performance_analysis}.

\paragraph{Subjective evaluation}
\begin{figure}[!t]
  \centering
  \includegraphics[width=\columnwidth]{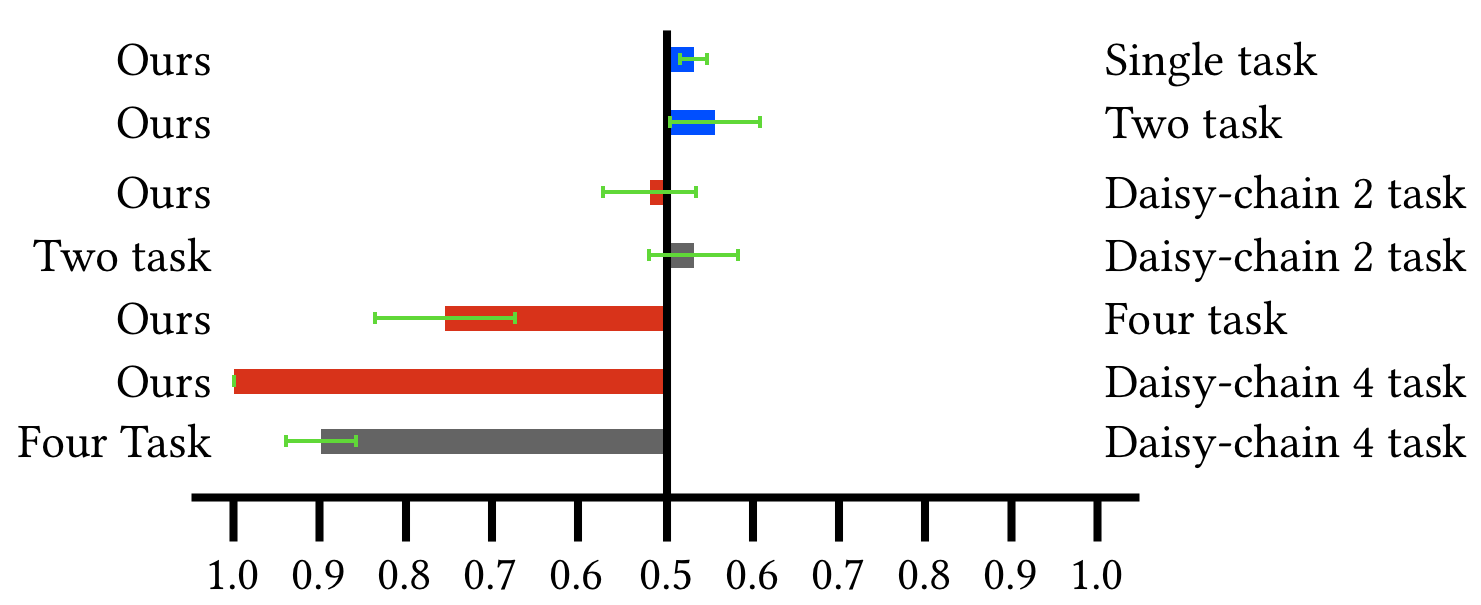}
  \caption{Preferences of participants in the user study.
  Our model is on-par with single-task models, two-task models, and daisy-chaining of two models, while outperforming the daisy-chaining of four tasks and the four-task model in terms of user preference. The colored bars indicate preference percentage, while green lines indicate a 95\% confidence interval.}
  \label{fig:user_study_results}
\end{figure}
We conducted an informal subjective study with 22 participants (age 18–30; 5 female, 17 male), all na\"ive to the study’s purpose.
The study comprised five sections, each assessing a different task using 15 image pairs.
Participant preferences between our model and the baselines are summarized in \refFig{user_study_results}, with 95\% confidence intervals.
Before each section, participants were informed about the task and asked to rate the image pairs based on their preferences.
Participants' preferences indicate that our model performs on par with single-task models, two-task models, and the daisy-chaining of two models, with half of the participants preferring our model.
For the four task cases, participants preferred our model with probabilities of 100\% and 74\% compared to the daisy-chaining of four tasks and the four-task model, respectively.
The participants' preferences are consistent with our quantitative results in \refTbl{quantitative_results} and further support our findings about complex tasks in \refSec{supporting_complex_tasks}.
Additional details about the user study are available in the supplementary's \refSec{user_study_details}.

\section{Discussion}
There are various limitations and potential future research directions that may help overcome the limitations in our learned multitasking perceptual graphics model.

\paragraph{Visual artifacts.}
In a small set of test cases where adjectives such as ``strongly,'' ``lightly,'' or ``mildly'' are used, we observe color deviations and a minor noise from the ground truth images.
In scenarios involving multiple tasks, ambiguous cases may occur when two effects target the same region of the image.
Task prioritization is crucial in such cases to avoid visual artifacts.
Our experiments indicate that our model tends to prioritize the task with the higher loss value, see supplementary's \refSec{task_prioritization_analysis}.
Among the tasks we support, chromostereopsis induces the largest change in pixel values and, as a result, is prioritized over other tasks.
For the extended discussion on visual artifacts, consult our supplementary's \refSec{visual_artifacts}.

\paragraph{Task-specific visual quality metrics}
Generic image quality metrics such as \PSNR, \SSIM, \LPIPS and \FovVideoVDP are not well suited for chromostereopsis and foveation cases, as shown in the first and third rows of \refFig{results0}.
In the case of foveation, these metrics fail to capture the metameric patterns in the peripheral regions, which are essential for the task.
In the case of chromostereopsis, they do not detect some artifacts that are present.
This limitation restricts the ability to further improve the quality of the generated images using learned methods, such as super-resolution \cite{lim2017enhanced}, as demonstrated in our supplementary \refSec{super_resolution_experiment}.

\paragraph{Supporting more tasks.}
Our model can potentially serve as a post-processor for many real-world applications, such as video streaming, image enhancement, and immersive display technologies.
For instance, several color-based perceptual tasks could be incorporated to reduce power usage \cite{duinkharjav2022color}, provide stereo view synthesis from a single image, and undertake prescription correction \cite{guzel2023chromacorrect}, anaglyph rendering \cite{woods2010comparing}, and accommodative rendering \cite{cholewiak2017chromablur} in immersive displays.
An extended discussion on the tasks currently supported can be found in the supplementary material's \refSec{task_selection_applications}.

\paragraph{Generalizing to unseen prompts and images.}
Unlike other methods \cite{sun2021task}, which utilize fixed one-hot-encoded text embeddings, our model can generalize to unseen prompts describing the supported tasks, offering a more flexible and user-friendly operation.
Examples demonstrating our model’s ability to generalize to unseen real-world images are provided in \refFig{results0}, and the supplementary's \refSec{out_of_dataset_generalization}.
Examples of prompt generalization are also available in the supplementary's \refSec{out_of_dataset_generalization}.
From these experiments, we can conclude that our model can generalize to unseen images and prompts that describe the supported tasks using novel wording.
When we use a prompt that describes an unsupported task, the resulting output image shows negligible changes, validating the language conditioning capability.

\paragraph{Conclusion}
The key finding from our model is effectively encoding images and text prompts for various perceptual tasks via multitask learning without exhausting computational resources.
With the help of this key finding, our model efficiently enhances images for immersive displays including \VR headsets and \AR glasses.

\section*{Acknowledgements}
The authors wish to thank Furkan Kınlı, Yicheng Zhan, Josef Spjut, and Morgan McGuire for fruitful discussions related to streaming and text-guidance aspects.
The authors thank anonymous reviewers for their feedback.

\bibliographystyle{ACM-Reference-Format}
\bibliography{sample-base}

%%% -*-BibTeX-*-
%%% Do NOT edit. File created by BibTeX with style
%%% ACM-Reference-Format-Journals [18-Jan-2012].

\begin{thebibliography}{46}

%%% ====================================================================
%%% NOTE TO THE USER: you can override these defaults by providing
%%% customized versions of any of these macros before the \bibliography
%%% command.  Each of them MUST provide its own final punctuation,
%%% except for \shownote{} and \showURL{}.  The latter two
%%% do not use final punctuation, in order to avoid confusing it with
%%% the Web address.
%%%
%%% To suppress output of a particular field, define its macro to expand
%%% to an empty string, or better, \unskip, like this:
%%%
%%% \newcommand{\showURL}[1]{\unskip}   % LaTeX syntax
%%%
%%% \def \showURL #1{\unskip}           % plain TeX syntax
%%%
%%% ====================================================================

\ifx \showCODEN    \undefined \def \showCODEN     #1{\unskip}     \fi
\ifx \showISBNx    \undefined \def \showISBNx     #1{\unskip}     \fi
\ifx \showISBNxiii \undefined \def \showISBNxiii  #1{\unskip}     \fi
\ifx \showISSN     \undefined \def \showISSN      #1{\unskip}     \fi
\ifx \showLCCN     \undefined \def \showLCCN      #1{\unskip}     \fi
\ifx \shownote     \undefined \def \shownote      #1{#1}          \fi
\ifx \showarticletitle \undefined \def \showarticletitle #1{#1}   \fi
\ifx \showURL      \undefined \def \showURL       {\relax}        \fi
% The following commands are used for tagged output and should be
% invisible to TeX
\providecommand\bibfield[2]{#2}
\providecommand\bibinfo[2]{#2}
\providecommand\natexlab[1]{#1}
\providecommand\showeprint[2][]{arXiv:#2}

\bibitem[Afifi and Brown(2020)]%
        {afifi2020deep}
\bibfield{author}{\bibinfo{person}{Mahmoud Afifi} {and}
  \bibinfo{person}{Michael~S Brown}.} \bibinfo{year}{2020}\natexlab{}.
\newblock \showarticletitle{Deep white-balance editing}. In
  \bibinfo{booktitle}{\emph{Proc. of the IEEE/CVF Conference on Computer Vision
  and Pattern Recognition}}. \bibinfo{pages}{1397--1406}.
\newblock


\bibitem[Bai et~al\mbox{.}(2016)]%
        {bai2016perceived}
\bibfield{author}{\bibinfo{person}{Yuejin Bai}, \bibinfo{person}{Yana Zhang},
  {and} \bibinfo{person}{Zhaohui Li}.} \bibinfo{year}{2016}\natexlab{}.
\newblock \showarticletitle{Perceived depth modeling based on
  chromostereopsis}. In \bibinfo{booktitle}{\emph{2016 2nd IEEE International
  Conference on Computer and Communications (ICCC)}}. IEEE,
  \bibinfo{pages}{723--727}.
\newblock


\bibitem[Brooks et~al\mbox{.}(2023)]%
        {brooks2023instructpix2pix}
\bibfield{author}{\bibinfo{person}{Tim Brooks}, \bibinfo{person}{Aleksander
  Holynski}, {and} \bibinfo{person}{Alexei~A Efros}.}
  \bibinfo{year}{2023}\natexlab{}.
\newblock \showarticletitle{Instructpix2pix: Learning to follow image editing
  instructions}. In \bibinfo{booktitle}{\emph{Proc. of the IEEE/CVF Conference
  on Computer Vision and Pattern Recognition}}. \bibinfo{pages}{18392--18402}.
\newblock


\bibitem[Caruana(1997)]%
        {caruana1997multitask}
\bibfield{author}{\bibinfo{person}{Rich Caruana}.}
  \bibinfo{year}{1997}\natexlab{}.
\newblock \showarticletitle{Multitask learning}.
\newblock \bibinfo{journal}{\emph{Machine learning}}  \bibinfo{volume}{28}
  (\bibinfo{year}{1997}), \bibinfo{pages}{41--75}.
\newblock


\bibitem[Choi and Han(2020)]%
        {choi2020task}
\bibfield{author}{\bibinfo{person}{Jinyoung Choi} {and}
  \bibinfo{person}{Bohyung Han}.} \bibinfo{year}{2020}\natexlab{}.
\newblock \showarticletitle{Task-aware quantization network for jpeg image
  compression}. In \bibinfo{booktitle}{\emph{Computer Vision--ECCV 2020: 16th
  European Conference, Glasgow, UK, August 23--28, 2020, Proceedings, Part XX
  16}}. Springer, \bibinfo{pages}{309--324}.
\newblock


\bibitem[Choi et~al\mbox{.}(2018)]%
        {Choi_2018_CVPR}
\bibfield{author}{\bibinfo{person}{Yunjey Choi}, \bibinfo{person}{Minje Choi},
  \bibinfo{person}{Munyoung Kim}, \bibinfo{person}{Jung-Woo Ha},
  \bibinfo{person}{Sunghun Kim}, {and} \bibinfo{person}{Jaegul Choo}.}
  \bibinfo{year}{2018}\natexlab{}.
\newblock \showarticletitle{StarGAN: Unified Generative Adversarial Networks
  for Multi-Domain Image-to-Image Translation}. In
  \bibinfo{booktitle}{\emph{Proc. of the IEEE Conference on Computer Vision and
  Pattern Recognition}}.
\newblock


\bibitem[Cholewiak et~al\mbox{.}(2017)]%
        {cholewiak2017chromablur}
\bibfield{author}{\bibinfo{person}{Steven~A Cholewiak},
  \bibinfo{person}{Gordon~D Love}, \bibinfo{person}{Pratul~P Srinivasan},
  \bibinfo{person}{Ren Ng}, {and} \bibinfo{person}{Martin~S Banks}.}
  \bibinfo{year}{2017}\natexlab{}.
\newblock \showarticletitle{Chromablur: Rendering chromatic eye aberration
  improves accommodation and realism}.
\newblock \bibinfo{journal}{\emph{ACM Transactions on Graphics (TOG)}}
  \bibinfo{volume}{36}, \bibinfo{number}{6} (\bibinfo{year}{2017}),
  \bibinfo{pages}{1--12}.
\newblock


\bibitem[Conde et~al\mbox{.}(2023)]%
        {conde2023perceptual}
\bibfield{author}{\bibinfo{person}{Marcos~V Conde}, \bibinfo{person}{Florin
  Vasluianu}, \bibinfo{person}{Javier Vazquez-Corral}, {and}
  \bibinfo{person}{Radu Timofte}.} \bibinfo{year}{2023}\natexlab{}.
\newblock \showarticletitle{Perceptual image enhancement for smartphone
  real-time applications}. In \bibinfo{booktitle}{\emph{Proc. of the IEEE/CVF
  Winter Conference on Applications of Computer Vision}}.
  \bibinfo{pages}{1848--1858}.
\newblock


\bibitem[Deza et~al\mbox{.}(2017)]%
        {deza2017towards}
\bibfield{author}{\bibinfo{person}{Arturo Deza}, \bibinfo{person}{Aditya
  Jonnalagadda}, {and} \bibinfo{person}{Miguel Eckstein}.}
  \bibinfo{year}{2017}\natexlab{}.
\newblock \showarticletitle{Towards metamerism via foveated style transfer}.
\newblock \bibinfo{journal}{\emph{arXiv preprint arXiv 1705.10041}}
  (\bibinfo{year}{2017}).
\newblock


\bibitem[Dong et~al\mbox{.}(2017)]%
        {dong2017semantic}
\bibfield{author}{\bibinfo{person}{Hao Dong}, \bibinfo{person}{Simiao Yu},
  \bibinfo{person}{Chao Wu}, {and} \bibinfo{person}{Yike Guo}.}
  \bibinfo{year}{2017}\natexlab{}.
\newblock \showarticletitle{Semantic image synthesis via adversarial learning}.
  In \bibinfo{booktitle}{\emph{Proc. of the IEEE International Conference on
  Computer Vision}}. \bibinfo{pages}{5706--5714}.
\newblock


\bibitem[Duinkharjav et~al\mbox{.}(2022)]%
        {duinkharjav2022color}
\bibfield{author}{\bibinfo{person}{Budmonde Duinkharjav},
  \bibinfo{person}{Kenneth Chen}, \bibinfo{person}{Abhishek Tyagi},
  \bibinfo{person}{Jiayi He}, \bibinfo{person}{Yuhao Zhu}, {and}
  \bibinfo{person}{Qi Sun}.} \bibinfo{year}{2022}\natexlab{}.
\newblock \showarticletitle{Color-perception-guided display power reduction for
  virtual reality}.
\newblock \bibinfo{journal}{\emph{ACM Transactions on Graphics (TOG)}}
  \bibinfo{volume}{41}, \bibinfo{number}{6} (\bibinfo{year}{2022}),
  \bibinfo{pages}{1--16}.
\newblock


\bibitem[Geng et~al\mbox{.}(2023)]%
        {Geng23instructdiff}
\bibfield{author}{\bibinfo{person}{Zigang Geng}, \bibinfo{person}{Binxin Yang},
  \bibinfo{person}{Tiankai Hang}, \bibinfo{person}{Chen Li},
  \bibinfo{person}{Shuyang Gu}, \bibinfo{person}{Ting Zhang},
  \bibinfo{person}{Jianmin Bao}, \bibinfo{person}{Zheng Zhang},
  \bibinfo{person}{Han Hu}, \bibinfo{person}{Dong Chen}, {and}
  \bibinfo{person}{Baining Guo}.} \bibinfo{year}{2023}\natexlab{}.
\newblock \showarticletitle{InstructDiffusion: {A} Generalist Modeling
  Interface for Vision Tasks}.
\newblock \bibinfo{journal}{\emph{CoRR}}  \bibinfo{volume}{abs/2309.03895}
  (\bibinfo{year}{2023}).
\newblock
\href{https://doi.org/10.48550/arXiv.2309.03895}{doi:\nolinkurl{10.48550/arXiv.2309.03895}}


\bibitem[G{\"u}zel et~al\mbox{.}(2023)]%
        {guzel2023chromacorrect}
\bibfield{author}{\bibinfo{person}{Ahmet~H G{\"u}zel}, \bibinfo{person}{Jeanne
  Beyazian}, \bibinfo{person}{Praneeth Chakravarthula}, {and}
  \bibinfo{person}{Kaan Akșit}.} \bibinfo{year}{2023}\natexlab{}.
\newblock \showarticletitle{ChromaCorrect: prescription correction in virtual
  reality headsets through perceptual guidance}.
\newblock \bibinfo{journal}{\emph{Biomedical Optics Express}}
  \bibinfo{volume}{14}, \bibinfo{number}{5} (\bibinfo{year}{2023}),
  \bibinfo{pages}{2166--2180}.
\newblock


\bibitem[Hong et~al\mbox{.}(2011)]%
        {hong2011depth}
\bibfield{author}{\bibinfo{person}{Ji~Young Hong}, \bibinfo{person}{Ho~Young
  Lee}, \bibinfo{person}{Du~Sik Park}, {and} \bibinfo{person}{Chang~Yeong
  Kim}.} \bibinfo{year}{2011}\natexlab{}.
\newblock \showarticletitle{Depth perception enhancement based on
  chromostereopsis}. In \bibinfo{booktitle}{\emph{Human Vision and Electronic
  Imaging XVI}}, Vol.~\bibinfo{volume}{7865}. SPIE, \bibinfo{pages}{367--376}.
\newblock


\bibitem[Huang et~al\mbox{.}(2023)]%
        {huang2023composer}
\bibfield{author}{\bibinfo{person}{Lianghua Huang}, \bibinfo{person}{Di Chen},
  \bibinfo{person}{Yu Liu}, \bibinfo{person}{Yujun Shen}, \bibinfo{person}{Deli
  Zhao}, {and} \bibinfo{person}{Jingren Zhou}.}
  \bibinfo{year}{2023}\natexlab{}.
\newblock \showarticletitle{Composer: Creative and controllable image synthesis
  with composable conditions}.
\newblock \bibinfo{journal}{\emph{arXiv preprint arXiv 2302.09778}}
  (\bibinfo{year}{2023}).
\newblock


\bibitem[Isola et~al\mbox{.}(2017)]%
        {isola2017pix2pix}
\bibfield{author}{\bibinfo{person}{Phillip Isola}, \bibinfo{person}{Jun-Yan
  Zhu}, \bibinfo{person}{Tinghui Zhou}, {and} \bibinfo{person}{Alexei~A
  Efros}.} \bibinfo{year}{2017}\natexlab{}.
\newblock \showarticletitle{Image-to-image translation with conditional
  adversarial networks}. In \bibinfo{booktitle}{\emph{Proc. of the IEEE
  conference on Computer Vision and Pattern Recognition}}.
  \bibinfo{pages}{1125--1134}.
\newblock


\bibitem[Jung and Ko(2012)]%
        {jung2012depth}
\bibfield{author}{\bibinfo{person}{Seung-Won Jung} {and}
  \bibinfo{person}{Sung-Jea Ko}.} \bibinfo{year}{2012}\natexlab{}.
\newblock \showarticletitle{Depth map based image enhancement using color
  stereopsis}.
\newblock \bibinfo{journal}{\emph{IEEE Signal Processing Letters}}
  \bibinfo{volume}{19}, \bibinfo{number}{5} (\bibinfo{year}{2012}),
  \bibinfo{pages}{303--306}.
\newblock


\bibitem[Karras et~al\mbox{.}(2019)]%
        {karras2019style}
\bibfield{author}{\bibinfo{person}{Tero Karras}, \bibinfo{person}{Samuli
  Laine}, {and} \bibinfo{person}{Timo Aila}.} \bibinfo{year}{2019}\natexlab{}.
\newblock \showarticletitle{A style-based generator architecture for generative
  adversarial networks}. In \bibinfo{booktitle}{\emph{Proc. of the IEEE/CVF
  conference on Computer Vision and Pattern Recognition}}.
  \bibinfo{pages}{4401--4410}.
\newblock


\bibitem[Ke et~al\mbox{.}(2023)]%
        {ke2023neural}
\bibfield{author}{\bibinfo{person}{Zhanghan Ke}, \bibinfo{person}{Yuhao Liu},
  \bibinfo{person}{Lei Zhu}, \bibinfo{person}{Nanxuan Zhao}, {and}
  \bibinfo{person}{Rynson~WH Lau}.} \bibinfo{year}{2023}\natexlab{}.
\newblock \showarticletitle{Neural preset for color style transfer}. In
  \bibinfo{booktitle}{\emph{Proc. of the IEEE/CVF Conference on Computer Vision
  and Pattern Recognition}}. \bibinfo{pages}{14173--14182}.
\newblock


\bibitem[Kim et~al\mbox{.}(2019)]%
        {kim2019foveated}
\bibfield{author}{\bibinfo{person}{Jonghyun Kim}, \bibinfo{person}{Youngmo
  Jeong}, \bibinfo{person}{Michael Stengel}, \bibinfo{person}{Kaan Aksit},
  \bibinfo{person}{Rachel~A Albert}, \bibinfo{person}{Ben Boudaoud},
  \bibinfo{person}{Trey Greer}, \bibinfo{person}{Joohwan Kim},
  \bibinfo{person}{Ward Lopes}, \bibinfo{person}{Zander Majercik},
  {et~al\mbox{.}}} \bibinfo{year}{2019}\natexlab{}.
\newblock \showarticletitle{Foveated AR: dynamically-foveated augmented reality
  display.}
\newblock \bibinfo{journal}{\emph{ACM Trans. Graph.}} \bibinfo{volume}{38},
  \bibinfo{number}{4} (\bibinfo{year}{2019}), \bibinfo{pages}{99--1}.
\newblock


\bibitem[Kirillov et~al\mbox{.}(2023)]%
        {kirillov2023segment}
\bibfield{author}{\bibinfo{person}{Alexander Kirillov}, \bibinfo{person}{Eric
  Mintun}, \bibinfo{person}{Nikhila Ravi}, \bibinfo{person}{Hanzi Mao},
  \bibinfo{person}{Chloe Rolland}, \bibinfo{person}{Laura Gustafson},
  \bibinfo{person}{Tete Xiao}, \bibinfo{person}{Spencer Whitehead},
  \bibinfo{person}{Alexander~C Berg}, \bibinfo{person}{Wan-Yen Lo},
  {et~al\mbox{.}}} \bibinfo{year}{2023}\natexlab{}.
\newblock \showarticletitle{Segment anything}.
\newblock \bibinfo{journal}{\emph{arXiv preprint arXiv 2304.02643}}
  (\bibinfo{year}{2023}).
\newblock


\bibitem[Ko et~al\mbox{.}(2023)]%
        {ko2023superstargan}
\bibfield{author}{\bibinfo{person}{Kanghyeok Ko}, \bibinfo{person}{Taesun
  Yeom}, {and} \bibinfo{person}{Minhyeok Lee}.}
  \bibinfo{year}{2023}\natexlab{}.
\newblock \showarticletitle{Superstargan: Generative adversarial networks for
  image-to-image translation in large-scale domains}.
\newblock \bibinfo{journal}{\emph{Neural Networks}}  \bibinfo{volume}{162}
  (\bibinfo{year}{2023}), \bibinfo{pages}{330--339}.
\newblock


\bibitem[Koulieris et~al\mbox{.}(2019)]%
        {koulieris2019near}
\bibfield{author}{\bibinfo{person}{George~Alex Koulieris},
  \bibinfo{person}{Kaan Ak{\c{s}}it}, \bibinfo{person}{Michael Stengel},
  \bibinfo{person}{Rafa{\l}~K Mantiuk}, \bibinfo{person}{Katerina Mania}, {and}
  \bibinfo{person}{Christian Richardt}.} \bibinfo{year}{2019}\natexlab{}.
\newblock \showarticletitle{Near-eye display and tracking technologies for
  virtual and augmented reality}. In \bibinfo{booktitle}{\emph{Computer
  Graphics Forum}}, Vol.~\bibinfo{volume}{38}. Wiley Online Library,
  \bibinfo{pages}{493--519}.
\newblock


\bibitem[Lim et~al\mbox{.}(2017)]%
        {lim2017enhanced}
\bibfield{author}{\bibinfo{person}{Bee Lim}, \bibinfo{person}{Sanghyun Son},
  \bibinfo{person}{Heewon Kim}, \bibinfo{person}{Seungjun Nah}, {and}
  \bibinfo{person}{Kyoung Mu~Lee}.} \bibinfo{year}{2017}\natexlab{}.
\newblock \showarticletitle{Enhanced deep residual networks for single image
  super-resolution}. In \bibinfo{booktitle}{\emph{Proc. of the IEEE conference
  on Computer Vision and Pattern Recognition workshops}}.
  \bibinfo{pages}{136--144}.
\newblock


\bibitem[Lin et~al\mbox{.}(2025)]%
        {lin2024pixwizard}
\bibfield{author}{\bibinfo{person}{Weifeng Lin}, \bibinfo{person}{Xinyu Wei},
  \bibinfo{person}{Renrui Zhang}, \bibinfo{person}{Le Zhuo},
  \bibinfo{person}{Shitian Zhao}, \bibinfo{person}{Siyuan Huang},
  \bibinfo{person}{Huan Teng}, \bibinfo{person}{Junlin Xie},
  \bibinfo{person}{Yu Qiao}, \bibinfo{person}{Peng Gao}, {et~al\mbox{.}}}
  \bibinfo{year}{2025}\natexlab{}.
\newblock \showarticletitle{Pixwizard: Versatile image-to-image visual
  assistant with open-language instructions}.
\newblock \bibinfo{journal}{\emph{Proc. of the International Conference on
  Learning Representations}} (\bibinfo{year}{2025}).
\newblock


\bibitem[Mantiuk et~al\mbox{.}(2021)]%
        {mantiuk2021fovvideovdp}
\bibfield{author}{\bibinfo{person}{Rafa{\l}~K Mantiuk}, \bibinfo{person}{Gyorgy
  Denes}, \bibinfo{person}{Alexandre Chapiro}, \bibinfo{person}{Anton
  Kaplanyan}, \bibinfo{person}{Gizem Rufo}, \bibinfo{person}{Romain Bachy},
  \bibinfo{person}{Trisha Lian}, {and} \bibinfo{person}{Anjul Patney}.}
  \bibinfo{year}{2021}\natexlab{}.
\newblock \showarticletitle{Fovvideovdp: A visible difference predictor for
  wide field-of-view video}.
\newblock \bibinfo{journal}{\emph{ACM Transactions on Graphics (TOG)}}
  \bibinfo{volume}{40}, \bibinfo{number}{4} (\bibinfo{year}{2021}),
  \bibinfo{pages}{1--19}.
\newblock


\bibitem[Marzuki and Sim(2020)]%
        {marzuki2020perceptual}
\bibfield{author}{\bibinfo{person}{Ismail Marzuki} {and}
  \bibinfo{person}{Donggyu Sim}.} \bibinfo{year}{2020}\natexlab{}.
\newblock \showarticletitle{Perceptual adaptive quantization parameter
  selection using deep convolutional features for HEVC encoder}.
\newblock \bibinfo{journal}{\emph{IEEE Access}}  \bibinfo{volume}{8}
  (\bibinfo{year}{2020}), \bibinfo{pages}{37052--37065}.
\newblock


\bibitem[Meng et~al\mbox{.}(2018)]%
        {10.1145/3203199}
\bibfield{author}{\bibinfo{person}{Xiaoxu Meng}, \bibinfo{person}{Ruofei Du},
  \bibinfo{person}{Matthias Zwicker}, {and} \bibinfo{person}{Amitabh
  Varshney}.} \bibinfo{year}{2018}\natexlab{}.
\newblock \showarticletitle{Kernel Foveated Rendering}.
\newblock \bibinfo{journal}{\emph{Proc. ACM Comput. Graph. Interact. Tech.}}
  \bibinfo{volume}{1}, \bibinfo{number}{1}, Article \bibinfo{articleno}{5}
  (\bibinfo{date}{jul} \bibinfo{year}{2018}), \bibinfo{numpages}{20}~pages.
\newblock
\href{https://doi.org/10.1145/3203199}{doi:\nolinkurl{10.1145/3203199}}


\bibitem[Ohayon et~al\mbox{.}(2021)]%
        {ohayon2021high}
\bibfield{author}{\bibinfo{person}{Guy Ohayon}, \bibinfo{person}{Theo Adrai},
  \bibinfo{person}{Gregory Vaksman}, \bibinfo{person}{Michael Elad}, {and}
  \bibinfo{person}{Peyman Milanfar}.} \bibinfo{year}{2021}\natexlab{}.
\newblock \showarticletitle{High perceptual quality image denoising with a
  posterior sampling cgan}. In \bibinfo{booktitle}{\emph{Proc. of the IEEE/CVF
  International Conference on Computer Vision}}. \bibinfo{pages}{1805--1813}.
\newblock


\bibitem[Ozolinsh and Muizniece(2015)]%
        {10.3389/fpsyg.2015.00337}
\bibfield{author}{\bibinfo{person}{Maris Ozolinsh} {and}
  \bibinfo{person}{Kristine Muizniece}.} \bibinfo{year}{2015}\natexlab{}.
\newblock \showarticletitle{Color Difference Threshold of Chromostereopsis
  Induced by Flat Display Emission}.
\newblock \bibinfo{journal}{\emph{Frontiers in Psychology}}
  \bibinfo{volume}{6} (\bibinfo{year}{2015}).
\newblock
\showISSN{1664-1078}
\href{https://doi.org/10.3389/fpsyg.2015.00337}{doi:\nolinkurl{10.3389/fpsyg.2015.00337}}


\bibitem[Patashnik et~al\mbox{.}(2021)]%
        {patashnik2021styleclip}
\bibfield{author}{\bibinfo{person}{Or Patashnik}, \bibinfo{person}{Zongze Wu},
  \bibinfo{person}{Eli Shechtman}, \bibinfo{person}{Daniel Cohen-Or}, {and}
  \bibinfo{person}{Dani Lischinski}.} \bibinfo{year}{2021}\natexlab{}.
\newblock \showarticletitle{Styleclip: Text-driven manipulation of stylegan
  imagery}. In \bibinfo{booktitle}{\emph{Proc. of the IEEE/CVF International
  Conference on Computer Vision}}. \bibinfo{pages}{2085--2094}.
\newblock


\bibitem[Radford et~al\mbox{.}(2021)]%
        {radford2021learning}
\bibfield{author}{\bibinfo{person}{Alec Radford}, \bibinfo{person}{Jong~Wook
  Kim}, \bibinfo{person}{Chris Hallacy}, \bibinfo{person}{Aditya Ramesh},
  \bibinfo{person}{Gabriel Goh}, \bibinfo{person}{Sandhini Agarwal},
  \bibinfo{person}{Girish Sastry}, \bibinfo{person}{Amanda Askell},
  \bibinfo{person}{Pamela Mishkin}, \bibinfo{person}{Jack Clark},
  {et~al\mbox{.}}} \bibinfo{year}{2021}\natexlab{}.
\newblock \showarticletitle{Learning transferable visual models from natural
  language supervision}. In \bibinfo{booktitle}{\emph{International conference
  on machine learning}}. PMLR, \bibinfo{pages}{8748--8763}.
\newblock


\bibitem[Ranftl et~al\mbox{.}(2021)]%
        {ranftl2021vision}
\bibfield{author}{\bibinfo{person}{Ren{\'e} Ranftl}, \bibinfo{person}{Alexey
  Bochkovskiy}, {and} \bibinfo{person}{Vladlen Koltun}.}
  \bibinfo{year}{2021}\natexlab{}.
\newblock \showarticletitle{Vision transformers for dense prediction}. In
  \bibinfo{booktitle}{\emph{Proc. of the IEEE/CVF International Conference on
  Computer Vision}}. \bibinfo{pages}{12179--12188}.
\newblock


\bibitem[Rombach et~al\mbox{.}(2022)]%
        {rombach2022high}
\bibfield{author}{\bibinfo{person}{Robin Rombach}, \bibinfo{person}{Andreas
  Blattmann}, \bibinfo{person}{Dominik Lorenz}, \bibinfo{person}{Patrick
  Esser}, {and} \bibinfo{person}{Bj{\"o}rn Ommer}.}
  \bibinfo{year}{2022}\natexlab{}.
\newblock \showarticletitle{High-resolution image synthesis with latent
  diffusion models}. In \bibinfo{booktitle}{\emph{Proc. of the IEEE/CVF
  conference on Computer Vision and Pattern Recognition}}.
  \bibinfo{pages}{10684--10695}.
\newblock


\bibitem[Sun et~al\mbox{.}(2021)]%
        {sun2021task}
\bibfield{author}{\bibinfo{person}{Guolei Sun}, \bibinfo{person}{Thomas
  Probst}, \bibinfo{person}{Danda~Pani Paudel}, \bibinfo{person}{Nikola
  Popovi{\'c}}, \bibinfo{person}{Menelaos Kanakis}, \bibinfo{person}{Jagruti
  Patel}, \bibinfo{person}{Dengxin Dai}, {and} \bibinfo{person}{Luc Van~Gool}.}
  \bibinfo{year}{2021}\natexlab{}.
\newblock \showarticletitle{Task switching network for multi-task learning}. In
  \bibinfo{booktitle}{\emph{Proc. of the IEEE/CVF International Conference on
  Computer Vision}}. \bibinfo{pages}{8291--8300}.
\newblock


\bibitem[Sun et~al\mbox{.}(2020)]%
        {sun2020adashare}
\bibfield{author}{\bibinfo{person}{Ximeng Sun}, \bibinfo{person}{Rameswar
  Panda}, \bibinfo{person}{Rogerio Feris}, {and} \bibinfo{person}{Kate
  Saenko}.} \bibinfo{year}{2020}\natexlab{}.
\newblock \showarticletitle{Adashare: Learning what to share for efficient deep
  multi-task learning}.
\newblock \bibinfo{journal}{\emph{Advances in Neural Information Processing
  Systems}}  \bibinfo{volume}{33} (\bibinfo{year}{2020}),
  \bibinfo{pages}{8728--8740}.
\newblock


\bibitem[Tariq and Didyk(2024)]%
        {tariq2024motion}
\bibfield{author}{\bibinfo{person}{Taimoor Tariq} {and} \bibinfo{person}{Piotr
  Didyk}.} \bibinfo{year}{2024}\natexlab{}.
\newblock \showarticletitle{Towards Motion Metamers for Foveated Rendering}.
\newblock \bibinfo{journal}{\emph{ACM Trans. Graph.}} (\bibinfo{year}{2024}).
\newblock


\bibitem[Vo and Sugimoto(2022)]%
        {vo2022paired}
\bibfield{author}{\bibinfo{person}{Duc~Minh Vo} {and} \bibinfo{person}{Akihiro
  Sugimoto}.} \bibinfo{year}{2022}\natexlab{}.
\newblock \showarticletitle{Paired-D++ GAN for image manipulation with text}.
\newblock \bibinfo{journal}{\emph{Machine Vision and Applications}}
  \bibinfo{volume}{33}, \bibinfo{number}{3} (\bibinfo{year}{2022}),
  \bibinfo{pages}{45}.
\newblock


\bibitem[Walton et~al\mbox{.}(2021)]%
        {walton2021beyond}
\bibfield{author}{\bibinfo{person}{David~R Walton},
  \bibinfo{person}{Rafael~Kuffner Dos~Anjos}, \bibinfo{person}{Sebastian
  Friston}, \bibinfo{person}{David Swapp}, \bibinfo{person}{Kaan Ak{\c{s}}it},
  \bibinfo{person}{Anthony Steed}, {and} \bibinfo{person}{Tobias Ritschel}.}
  \bibinfo{year}{2021}\natexlab{}.
\newblock \showarticletitle{Beyond blur: Real-time ventral metamers for
  foveated rendering}.
\newblock \bibinfo{journal}{\emph{ACM Transactions on Graphics}}
  \bibinfo{volume}{40}, \bibinfo{number}{4} (\bibinfo{year}{2021}),
  \bibinfo{pages}{1--14}.
\newblock


\bibitem[Westermann(2022)]%
        {westermann2022using}
\bibfield{author}{\bibinfo{person}{Helena Westermann}.}
  \bibinfo{year}{2022}\natexlab{}.
\newblock \showarticletitle{Using Chromostereopsis to Enhance Depth Perception
  in Photos by changing the Hue}.
\newblock  (\bibinfo{year}{2022}).
\newblock


\bibitem[Woods and Harris(2010)]%
        {woods2010comparing}
\bibfield{author}{\bibinfo{person}{Andrew~J Woods} {and}
  \bibinfo{person}{Chris~R Harris}.} \bibinfo{year}{2010}\natexlab{}.
\newblock \showarticletitle{Comparing levels of crosstalk with red/cyan,
  blue/yellow, and green/magenta anaglyph 3D glasses}. In
  \bibinfo{booktitle}{\emph{Stereoscopic displays and applications XXI}},
  Vol.~\bibinfo{volume}{7524}. SPIE, \bibinfo{pages}{235--246}.
\newblock


\bibitem[Xia et~al\mbox{.}(2021)]%
        {xia2021tedigan}
\bibfield{author}{\bibinfo{person}{Weihao Xia}, \bibinfo{person}{Yujiu Yang},
  \bibinfo{person}{Jing-Hao Xue}, {and} \bibinfo{person}{Baoyuan Wu}.}
  \bibinfo{year}{2021}\natexlab{}.
\newblock \showarticletitle{Tedigan: Text-guided diverse face image generation
  and manipulation}. In \bibinfo{booktitle}{\emph{Proc. of the IEEE/CVF
  conference on Computer Vision and Pattern Recognition}}.
  \bibinfo{pages}{2256--2265}.
\newblock


\bibitem[Yu et~al\mbox{.}(2020)]%
        {yu2020low}
\bibfield{author}{\bibinfo{person}{Haibao Yu}, \bibinfo{person}{Tuopu Wen},
  \bibinfo{person}{Guangliang Cheng}, \bibinfo{person}{Jiankai Sun},
  \bibinfo{person}{Qi Han}, {and} \bibinfo{person}{Jianping Shi}.}
  \bibinfo{year}{2020}\natexlab{}.
\newblock \showarticletitle{Low-bit quantization needs good distribution}. In
  \bibinfo{booktitle}{\emph{Proc. of the IEEE/CVF Conference on Computer Vision
  and Pattern Recognition Workshops}}. \bibinfo{pages}{680--681}.
\newblock


\bibitem[Zhang et~al\mbox{.}(2017)]%
        {zhang2017beyond}
\bibfield{author}{\bibinfo{person}{Kai Zhang}, \bibinfo{person}{Wangmeng Zuo},
  \bibinfo{person}{Yunjin Chen}, \bibinfo{person}{Deyu Meng}, {and}
  \bibinfo{person}{Lei Zhang}.} \bibinfo{year}{2017}\natexlab{}.
\newblock \showarticletitle{Beyond a gaussian denoiser: Residual learning of
  deep cnn for image denoising}.
\newblock \bibinfo{journal}{\emph{IEEE transactions on image processing}}
  \bibinfo{volume}{26}, \bibinfo{number}{7} (\bibinfo{year}{2017}),
  \bibinfo{pages}{3142--3155}.
\newblock


\bibitem[Zhang et~al\mbox{.}(2018)]%
        {zhang2018unreasonable}
\bibfield{author}{\bibinfo{person}{Richard Zhang}, \bibinfo{person}{Phillip
  Isola}, \bibinfo{person}{Alexei~A Efros}, \bibinfo{person}{Eli Shechtman},
  {and} \bibinfo{person}{Oliver Wang}.} \bibinfo{year}{2018}\natexlab{}.
\newblock \showarticletitle{The unreasonable effectiveness of deep features as
  a perceptual metric}. In \bibinfo{booktitle}{\emph{Proc. of the IEEE
  conference on Computer Vision and Pattern Recognition}}.
  \bibinfo{pages}{586--595}.
\newblock


\bibitem[Zhu et~al\mbox{.}(2017)]%
        {zhu2017unpaired}
\bibfield{author}{\bibinfo{person}{Jun-Yan Zhu}, \bibinfo{person}{Taesung
  Park}, \bibinfo{person}{Phillip Isola}, {and} \bibinfo{person}{Alexei~A
  Efros}.} \bibinfo{year}{2017}\natexlab{}.
\newblock \showarticletitle{Unpaired image-to-image translation using
  cycle-consistent adversarial networks}. In \bibinfo{booktitle}{\emph{Proc. of
  the IEEE International Conference on Computer Vision}}.
  \bibinfo{pages}{2223--2232}.
\newblock


\end{thebibliography}

\end{document}